%% file: main.tex
\documentclass[twoside,11pt]{article}

\usepackage{blindtext}

\usepackage{tikz}
\usepackage[ruled,vlined,linesnumbered]{algorithm2e}
\usepackage{caption}
\usepackage{subcaption}

\usepackage{color}
\usepackage{latexsym}
\usepackage{amsmath}               
\usepackage{amssymb}               
\usepackage{amsfonts}              
\usepackage{multirow}
\usepackage{tikz}                  
\usetikzlibrary{arrows,positioning,shapes}
\usepackage{tcolorbox}
\usepackage{nicefrac}
\usepackage{dsfont}

\RequirePackage[%
  pdfstartview=FitH,%
  breaklinks=true,%
  bookmarks=true,%
  colorlinks=true,%
  linkcolor= blue,
  anchorcolor=blue,%
  citecolor=blue,
  filecolor=blue,%
  menucolor=blue,%
  urlcolor=blue%
  ]{hyperref}
  \usepackage{cleveref}
  \crefname{prop}{Proposition}{Propositions}
  \crefname{lemma}{Lemma}{Lemmas}
  \crefname{corollary}{Corollary}{Corollaries}
  \crefname{example}{Example}{Examples}
  \crefname{appendix}{Appendix}{Appendixes}
  \crefname{remark}{Remark}{Remark}
  \crefname{figure}{Figure}{Figures}

\usepackage[nohyperref,preprint]{jmlr2e}
\usepackage{lastpage}
\jmlrheading{23}{2023}{1-\pageref{LastPage}}{5/23}{}{23-0000}{Maxime Haddouche and Benjamin Guedj}

\ShortHeadings{Wasserstein PAC-Bayes Learning}{Haddouche and Guedj}
\firstpageno{1}

\begin{document}

\title{Wasserstein PAC-Bayes Learning: Exploiting Optimisation Guarantees to Explain Generalisation}

\author{\name Maxime Haddouche \email maxime.haddouche@inria.fr \\
       \addr Inria, University College London\\
        and Université de Lille\\
       France
       \AND
       \name Benjamin Guedj \email benjamin.guedj@inria.fr \\
       \addr Inria and University College London \\
       UK
       }

\editor{}

\maketitle

\begin{abstract}
PAC-Bayes learning is an established framework to both assess the generalisation ability of learning algorithms, and design new learning algorithm by exploiting generalisation bounds as training objectives. Most of the exisiting bounds involve a \emph{Kullback-Leibler} (KL) divergence, which fails to capture the geometric properties of the loss function which are often useful in optimisation. We address this by extending the emerging \emph{Wasserstein PAC-Bayes} theory. We develop new PAC-Bayes bounds with Wasserstein distances replacing the usual KL, and demonstrate that sound optimisation guarantees translate to good generalisation abilities. In particular we provide generalisation bounds for the \emph{Bures-Wasserstein SGD} by exploiting its optimisation properties.
\end{abstract}

\begin{keywords}
  PAC-Bayes, Wasserstein distance, Generalisation, Optimisation
\end{keywords}


\input{body}


\acks{
B.G. acknowledges partial support by the U.S. Army Research Laboratory and the U.S. Army Research Office, and by the U.K. Ministry of Defence and the U.K. Engineering and Physical Sciences Research Council (EPSRC) under grant number EP/R013616/1. B.G. acknowledges partial support from the French National Agency for Research, grants ANR- 18-CE40-0016-01 and ANR-18-CE23-0015-02.}

\newpage

\appendix

\input{appendix}

\bibliography{biblio}

\end{document}

%% file: body.tex

\section{Introduction and state-of-the-art results}

\textit{On PAC-Bayes learning.}
PAC-Bayes (see the seminal works of \citealp{STW1997},\citealp{McAllester1998,McAllester1999,McAllester2003} and \citealp{catoni2003pac,catoni2007pac},
see also \cite{guedj2019primer,alquier2021survey} for recent surveys) is a powerful framework to explain the generalisation ability of learning algorithms, in the sense that it provides an upper bound on the generalisation gap.
Indeed, PAC-Bayes theory aims to upper bound the gap between the averaged error on a novel unseen datum and the empirical performance on a training set, without involving test data.
PAC-Bayes guarantees consist in empirical upper bounds, obtained through various techniques such as exponential moments of a Bernoulli \citep{McAllester2003} the log-Laplace transform \citep{catoni2007pac}, Bernstein inequality \citep{tolstikhin2013pac, mhammedi2019} among others.
This is crucial as it gives PAC-Bayes a wider range: beyond guarantees for existing algorithms, PAC-Bayes bounds lead to novel learning procedures by defining new training objectives.
Such algorithms have been instantiated in several learning problems, \emph{e.g.}, deep nets \citep{dziugaite2017computing,letarte2019dichotomize,rivasplata2019pac,perez2021tighter,biggs2021differentiable,perezortiz2021learning,perezortiz2021progress,biggs2022non},
meta-learning \citep{amit2018meta,FaridMajumdar2021,rothfuss2021pacoh,DingChenLevinboimGoodmanSoricut2021,rothfuss2022pac,riou2023bayes},
online learning \citep{haddouche2022online}, reinforcement learning \citep{fard2010pac} or bandits \citep{seldin2011pac}.

\textit{A theory built around the KL divergence.}
Most of the PAC-Bayes literature is based on the use of the \emph{Kullback-Leibler divergence} (KL) (and more recently $f$-divergences, \citealp{ohnishi2021novel,picard2022divergence}) to control the discrepancy of a \emph{posterior} distribution (over the predictor space of interest) from a \emph{prior} distribution, the posterior being data-dependent while the prior is data-independent. However, the KL divergence suffers from limitations as it does not satisfy classical properties such as the triangle inequality or even symmetry: it is challenging to exploit geometric properties of the measure space and the loss function through it.

\textit{PAC-Bayes learning with Wasserstein distances.}
A recent line of work led by \citet{amit2022ipm} investigates PAC-Bayes generalisation bounds with a Wasserstein distance rather than the KL. This idea has been simultaneously developed by \citet{ohana2022shedding} for sliced adaptive Wasserstein distances.
Also the recent work of \citet{mbacke2023pacbayesian} provides PAC-Bayesian bounds for adversarial generative models where the quantity of interest is a Wasserstein distance (although the complexity measure remains a KL divergence).

In the present paper, we propose a major development of the emerging \emph{Wasserstein PAC-Bayes} (WPB) theory.
\citet{amit2022ipm} provided the first high-probability WPB bounds with explicit convergence rates (for bounded losses) only for finite predictor classes or for linear regression problems. We extend those results to a broader framework including uncountable predictor classes and unbounded losses. We first propose a novel  WPB bound valid on any compact for bounded lipschitz losses. From this, we demonstrate that the WPB framework allows to bypass both the compactness assumption on the predictor class and the bounded loss assumption: Wasserstein PAC-Bayes only requires Lipschitz or smooth functions to be used. We obtain explicit bounds for the case of prior and posterior distributions taken within a compact space of Gaussian measures.
We also extend those results to the case of data-dependent priors, which is of interest when one compares the output of an algorithmic procedure to its minimisation objective.

As Wasserstein distance recently appeared as complexity measure in expected generalisation bounds (see \emph{e.g.} \citealp{rodriguez2021tight}), the high-probability Wasserstein PAC-Bayes bounds presented here investigate deeper this lead. We also go a step further by showing that Wasserstein PAC-Bayes allows to reap the benefits of optimisation guarantees within generalisation. To the best of our knowledge, no previous PAC-Bayes bound has achieved this goal.
More precisely, we focus on the Bures-Wasserstein SGD \citep{altschuler2021aver,lambert2022variational} and show that the output of this algorithm, with enough data, after enough optimisation steps, is able to generalise well, independently of the quality of the initialisation point. The take-home message is that if an optimisation method has convergence guarantees with respect to a Wasserstein distance, then WPB theory allow us to determine, before any training, whether the algorithmic output will generalise well.

\textit{Outline.}
The remainder of this section is structured as follows: we state in \Cref{sec: framework} the framework and notation. In \Cref{sec: intro_optim}, we describe how current PAC-Bayes procedures are designed and how their efficiency is evaluated, and we discuss current limitations.
In \Cref{sec: intro_contrib}, we describe our main contributions, showing how we establish a WPB theory (using techniques which differ from those in \citealp{amit2022ipm}) in order to exploit the optimisation results of \citet{lambert2022variational}.

\Cref{sec: compact_space} gathers results for compact predictor spaces, \Cref{sec: wpb_gauss} gives WPB bounds for Gaussian prior and posterior, \Cref{sec: data_dep_priors} contains a WPB bound with a data-dependent prior for unbounded Lipschitz losses. \Cref{sec: gene_sgd} establishes a link between optimisation and generalisation by exploiting the results of \citet{lambert2022variational} to establish new generalisation guarantees for the Bures-Wasserstein SGD. We defer to \Cref{sec: background} additional background notes and to \Cref{sec: proofs} proofs which are not essential to the understanding of our contributions.

\subsection{Framework}
\label{sec: framework}

\textit{Learning theory framework.}
We consider a \emph{learning problem} specified by a tuple $(\mathcal{H}, \mathcal{Z}, \ell)$ of a set $\mathcal{H}$ of predictors, a data space $\mathcal{Z}$, and a loss function $\ell : \mathcal{H}\times \mathcal{Z} \rightarrow \mathbb{R} $.
We consider a finite dataset $S=(z_i)_{i\in \{1..m\}}\in\mathcal{Z}^{m}$ and assume that sequence is i.i.d. following the distribution $\mu$. We always assume that $\mathcal{H}\subseteq\mathbb{R}^d$, we denote by $\Sigma_{\mathcal{H}}$ the associated Borel $\sigma$-algebra and we denote by $||.||$ the classical Euclidean norm. We denote by $\mathcal{M}_1(\mathcal{H})$ the set of probability measures on $\mathcal{H}$.
We denote by $\mathcal{P}_1(\mathcal{H})$ (resp. $\mathcal{P}_2(\mathcal{H})$) the subspace of $\mathcal{M}_1(\mathcal{H})$ of with finite order 1 (resp. order 2) moments wrt $||.||$.

\textit{Definitions.}
The \emph{generalisation error} $R$ of any predictor $h\in\mathcal{H}$ is $R(h)= \mathbb{E}_{z\sim \mu}[\ell(h,z)]$, the \emph{empirical error} of $h$ is $R_S(h)= \frac{1}{m}\sum_{i=1}^m\ell(h,z_i)]$.
The \emph{generalisation gap} of any $h$ is the quantity $\Delta_S(h)=R(h)-R_S(h)$ and, for any $Q\in\mathcal{M}_1(H)$, $\Delta_S(Q)= \mathbb{E}_{h\sim Q}[\Delta_S(h)]$. In what follows, we let $\mathcal{B}(x,r)$ (resp. $\bar{\mathcal{B}}(x,r)$) denote the ball (resp. closed ball) centered in $x\in\mathbb{R}^d$ of radius $r$.
We define the \emph{Gibbs posterior} associated to the prior $P\in\mathcal{M}_1(\mathcal{H})$ as the measure $P_{-\lambda R_S}$ such that $\mathrm{d}P_{-\lambda R_S} \propto \exp(-\lambda R_S(.)) \mathrm{d}P(.)$.

We denote by $\text{BW}(\mathbb{R}^d)\subset \mathcal{P}_2(\mathbb{R}^d)$ the set of non-degenerate Gaussian distributions, also known as the \emph{Bures-Wasserstein space}. For a measurable function $T:\mathbb{R}^d \rightarrow \mathbb{R}^d$, and a measure $P\in\mathcal{P}_1(\mathbb{R}^d)$ we let $T\#P$ denote the measure such that for any $B\in \Sigma_{\mathbb{R}^d}, T\#P(B)= P(T^{-1}(B))$.
For any $R>0$, we denote by $\mathcal{P}_R$ the projection over $\bar{\mathcal{B}}(0_{\mathbb{R}^d},R)$.
Finally, as we consider compact sets of $\text{BW}(\mathbb{R}^d)$, we define for any $0\leq \alpha\leq \beta, M\geq 0$ the set
\[ C_{\alpha,\beta,M} := \left\{ \mathcal{N}(m,\Sigma) \in \text{BW}(\mathbb{R}^d) \mid ||m||\leq M,\; \alpha \mathrm{Id} \preceq \Sigma \preceq \beta \mathrm{Id} \right\}.  \]

\subsection{PAC-Bayes and optimisation: limits and caveats}
\label{sec: intro_optim}
\textit{Optimisation in PAC-Bayes.}
PAC-Bayesian generalisation bounds are meant to control how well measures derived from a learning algorithm perform on novel data.  Those bounds involves a complexity term which is typically a Kullback Leibler (KL) divergence. A prototypic bound is as follows: with probability $1-\delta$, for all measure $Q$,

$$\Delta_S(Q) \leq \sqrt{\frac{\operatorname{COMP}(Q)}{m}}, $$
where $\operatorname{COMP}$ is a complexity term involving a data-free prior $P$ and an approximation term $1-\delta$.
From an optimisation perspective, this upper bound can be seen as a learning objective, where $\operatorname{COMP}$ acts as a regulariser to avoid overfitting on the empirical risk:

$$ Q^* := \underset{Q\in\mathcal{M}_1(\mathcal{H})}{\operatorname{argmin}} R_S(Q) + \sqrt{\frac{\operatorname{COMP}(Q)}{m}}.$$
Such algorithms are build to ensure a candidate measure with a good generalisation ability.
However the convergence of the optimisation process remains unclear: as $\sqrt{\operatorname{COMP}}$ is not necessarily convex in $Q$, it is unclear whether an optimisation procedure on the previous learning objective will lead to $\hat{Q}$ (or a good approximation of it). A good introductory example is to optimise the PAC-Bayesian learning objective for the following complexity term, holding for a loss $\ell$ being in $[0,1]$:

$$ \sqrt{\frac{\operatorname{COMP}(Q)}{m}} := \frac{\operatorname{KL}(Q,P)}{\lambda} + \frac{\lambda}{2m},$$
with $\lambda$ being usually fine-tuned over a countable grid.
This objective, linear in the KL divergence term is optimised by the Gibbs posterior:
$$ \mathrm{d}Q^*(h)\propto \exp(-\lambda R_S(h)) \mathrm{d}P(h).$$
This distribution, while being known analytically, may be hard to compute in practice. A class of methods dedicated to compute or approximate this posterior distribution are the Markov Chain Monte Carlo (MCMC) methods that rely on carefully constructed Markov chains which (approximately) converge to $Q^*$. However, MCMC methods can be computationally costly and other methods were studied to obtain quickly surrogates of $Q^*$. In particular, \emph{Variational Inference} (VI) has been developed as a time-efficient solution. VI algorithms aims to estimate a surrogate $\hat{Q}$ of $Q^*$, often chosen within a parametric class of measures such as Gaussian measures. For instance, in order to approximate $Q^*$ it is natural to consider the following surrogate:
$$ \hat{Q} = \underset{Q\in \mathcal{C}}{\operatorname{argmin}} \operatorname{KL}(Q,Q^*) ,$$
where $\mathcal{C}$ is a subset of $\mathcal{M}_1(\mathcal{H})$. When $\mathcal{C}$ is the set of Gaussian measures (also known as the \emph{Bures-Wasserstein} manifold), the convergence of the associated VI algorithm has been studied \citep{altschuler2021aver,lambert2022variational}.
This candidate $\hat{Q}$ is approximated after $N$ optimisation steps by a measure $\hat{Q}_N$ and is then used in McAllester's bound to assess its efficiency:
\begin{align}
\label{eq: eval_posterior}
\Delta_S(\hat{Q}_N) \leq \sqrt{\frac{\operatorname{KL}(\hat{Q}_N,P) + \log(m/\delta)}{2m}}.
\end{align}

\textit{Role of the prior $P.$}
From an optimisation perspective, the conclusion of \eqref{eq: eval_posterior} is that if $\hat{Q}_N$ is a good approximation of $\hat{Q}$ and if the initialisation $P$ is well-chosen, then the generalisation ability $\hat{Q}$ is guaranteed to be high. Assuming such a condition on $P$ may be unrealistic. Furthermore the term $\operatorname{KL}(\hat{Q}_N,P)$ acts as a blackbox as we do not have a theoretical control on how far $\hat{Q}$ and $\hat{Q}_N$ diverge from the prior.
In particular if the prior is ill-chosen, then we could have $\operatorname{KL}(\hat{Q}_N,P) = \mathcal{O}(m)$, making \eqref{eq: eval_posterior} vacuous.

\textit{Data-dependent priors are not enough to explain the generalisation gain through optimisation.}
As shown above, in order to have a sound theoretical control on the generalisation ability of the algorithmic output $\hat{Q}_N$, it is irrelevant to compare it to the initialisation $P$. Thus, it is legitimate to wonder if the existing PAC-Bayesian techniques using data-dependent priors are enough to fill this gap. To do so, we identify two strategies.
\begin{enumerate}
  \item Taking $Q^*$ as a 'prior' distribution (as advised by \citealp{dziugaite2017computing}) is, at first sight, a convincing answer. However, the use of KL divergence is problematic. Indeed, we cannot make $\hat{Q}$ appear easily in \Cref{eq: eval_posterior} which is the relevant point of interest. Furthermore, to our knowledge, there is no VI algorithm which guarantees that $\operatorname{KL}(\hat{Q}_N,Q^*)$ is decreasing.
  \item The prior is obtained from an algorithmic method on a fraction of training data. Then, such a bound does not inform us whether the considered optimisation method has been able to reach an optimum during the training phase: similarly to a test bound, it mainly assesses the post-training efficiency of the output of the learning algorithm. A relevant example is Table 3 of \citet{perezortiz2021learning} which considers data-dependent priors obtained through SGD. Then as the performance of the prior and the posterior is roughly similar, it is hard to determine whether the associated theoretical guarantee is more meaningful than a test bound as the prior measure could have already converged near a local optimum.
\end{enumerate}

\textit{A strategy to replace \eqref{eq: eval_posterior}.}
In order to assess whether the output of a learning algorithm enjoys high generalisation, a PAC-Bayes bound should satisfy the following generic form:
\begin{align}
\label{eq: wanted_pattern}
\Delta_S(\hat{Q}_N) \leq \sqrt{\frac{f(N)\operatorname{D}(P,\hat{Q}) + \varepsilon+ \log(m/\delta)}{2m}},
\end{align}
where $f$ is a function decreasing to $0$ as $N$ goes to infinity, which comes from the optimisation procedure, $\operatorname{D}$ is the way to measure the discrepancy between $P,\hat{Q}$ (classically it would be the KL divergence) and $\varepsilon$ is a residual term which could contain for instance the discrepancy $\operatorname{KL}(Q^*,\hat{Q})$ between the approximation and the true minimiser.
Such a guarantee would give theoretical evidence that the generalisation ability of $\hat{Q}_N$ is independent of the choice of the initialisation point $P$ and tends to $\mathcal{O}\left( \sqrt{\frac{\varepsilon + \log(m/\delta)}{m}} \right)$.
To the best of our knowledge, there is no work proposing an optimisation procedure such that $\operatorname{KL}(\hat{Q}_N,\hat{Q}) \leq f(N)  \operatorname{KL}(P,\hat{Q})$.  This lack is unfortunate but not surprising as the $KL$ divergence is not a distance: it is not easy to incorporate optimisation guarantees, often based on geometric properties of the loss, into the KL divergence.

\textit{Our aims in this paper.}
A legitimate question is then: is it possible to extend the PAC-Bayes theory beyond the KL divergence in order to explain before training, with a bound of the form of \eqref{eq: wanted_pattern}, whether the output of optimisation procedure have high generalisation ability? We structure the present paper to provide a positive answer to this question. More precisely we develop a WPB bound of the form of \eqref{eq: wanted_pattern} for the output of the Bures-Wasserstein SGD \citep{lambert2022variational}.

\subsection{Summary of our contributions}
\label{sec: intro_contrib}
To make PAC-Bayes learning useful to explain the generalisation ability of minimisers reached by optimisation algorithms, we develop theoretical results built around Wasserstein distances whose definitions are recalled below.

\begin{definition}
\label{def: wasserstein}
The $1$-Wasserstein distance between $P,Q \in \mathcal{P}_1(\mathcal{H})$ is defined as
\[ W_1(Q,P) = \inf_{\pi \in \Pi(Q,P)} \int_{\mathcal{H}^2} ||x-y||\mathrm{d}\pi(x,y). \]
where $\Pi(Q,P)$ denote the set of probability measures on $\mathcal{H}^2$ whose marginals are $Q$ and $P$.
We define the $2$-Wasserstein distance on $\mathcal{P}(\mathcal{H})$ as
\[ W_2(Q,P) = \sqrt{\inf_{\pi \in \Pi(Q,P)} \int_{\mathcal{H}^2} ||x-y||^2\mathrm{d}\pi(x,y)}. \]
\end{definition}
\citet{amit2022ipm} provided a preliminary WPB bound, being explicit for the case of finite predictor classes and linear regression problems. To do so, they exploited the Kantorovich-Rubinstein duality (see, \emph{e.g.}, Remark 6.5 in \citealp{villani2009optimal}) of the $1$-Wasserstein distance. We exploit another duality formula (Theorem 5.10 in \citealp{villani2009optimal}) valid for any cost function (in the framework of optimal transport). This leads to a WPB bound valid for \emph{uniformly Lipschitz} loss functions.
\begin{definition}
\label{def: unif_lpz}
We say that a function $\ell:\mathcal{H}\times\mathcal{Z}\rightarrow \mathbb{R}$ is \emph{uniformly $K$-Lipschitz} if for any $z\in\mathcal{Z}$, $\ell(.,z)$ is $K$-Lipschtiz. We also say that a function is \emph{uniformly L-smooth} (or simply smooth) if for any $z\in\mathcal{Z}$, its gradient $\nabla \ell(.,z)$ is $L$-Lipschitz.
\end{definition}

\textit{A WPB bound for compact predictor classes.}
We first extend the PAC-Bayes framework to the case where the discrepancy between measures is expressed through the $1$-Wasserstein distance. It is stated as follows: for uniformly $K$-lipschitz functions bounded in $[0,1]$ with  $\mathcal{H}\subseteq \mathcal{B}_R:= \bar{\mathcal{B}}(0_{\mathbb{R}^d},R)$, we have for any prior $P\in\mathcal{M}_1(\mathcal{H})$, with probability at least $1-\delta$, for any posterior distribution $Q\in\mathcal{M}_1(\mathcal{H})$
\[ |\Delta_S(Q)| \leq \mathcal{O}\left(\sqrt{2K(2K+1)\frac{2d\log\left(3\frac{1 +2Rm }{\delta}\right)}{m} \left(1+W_1(Q,P)  \right) +\frac{\log\left( \frac{m}{\delta} \right)}{m} } \right). \]
This bound extends the WPB bound of \citet{amit2022ipm} to the case of a compact space of predictors. The proof technique exploits covering number arguments to prove the Lipschitzness (with high probability) of a relevant functional. The duality theorem of \citet[Theorem 5.10]{villani2009optimal} allows us to generate a local change of measure inequality (see, \emph{e.g.}, \citealp{donsker1975asymptotic}) required to use PAC-Bayes learning.
This bound is stated in \Cref{th: compact_mcall} and further discussed in \Cref{sec: compact_space}. However, this result does not cover the celebrated case of PAC-Bayes with Gaussian priors and posteriors. We then develop the next result to address this important case.

\textit{WPB bounds with Gaussians measures for unbounded losses.} Through the calculus of the residuals of Euler's Gamma function we obtain in \Cref{th: main_gaussian_lpz}, stated in \Cref{sec: wpb_gauss}, the following result when $\mathcal{H}=\mathbb{R}^d$, for loss functions lying in $[0,1]$ being uniformly $K$-lipschitz: for any gaussian prior $P$ in a compact $C_{\alpha,\beta,M}\subseteq \operatorname{BW}(\mathbb{R}^d)$, with probability at least $1-\delta$, for any posterior distribution $Q\in \mathcal{C}$,
\[ |\Delta_S(Q)| \leq \mathcal{O}\left(\sqrt{2K(2K+1)\frac{2d\log\left(3\frac{1 +2Rm }{\delta}\right)}{m} \left(1+ \sqrt{\nicefrac{d}{m}} + W_1(Q,P)  \right) +\frac{\log\left( \frac{m}{\delta} \right)}{m}} \right), \]
where $R= \mathcal{O}(\max \sqrt{d\log(d)},\sqrt{\log(m)})$.
This shows that, using $R$ as an hyperparameter, we are able to maintain nearly the same convergence rate than \Cref{th: compact_mcall} at the cost of an extra factor of $\sqrt{\log(dm)}$.
Interestingly, we are able to remove in \Cref{cor: unbounded_lpz} the boundedness assumption to obtain a WPB bound, valid for unbounded uniformly $K$-lipschitz function  with an additional boundedness assumption on $\sup_{z} \ell(0,z)$. This bound is more sensitive to the dimension of the problem when few data points are available. However, the asymptotic dependency remains (nearly) unchanged, at the cost of an extra polynomial factor in $\log(dm)$:
\begin{align}
|\Delta_S(Q)|  \leq\Tilde{\mathcal{O}}\left( \sqrt{2K\frac{d}{m}\left(1 + W_1(Q,P)\right)+(1+K^2\log(m))\frac{\log\left( \frac{m}{\delta} \right)}{m}}   \right).
\end{align}
$\Tilde{\mathcal{O}}$ hides a polynomial dependency in $(\log(d),\log(m))$. This result is further discussed ion \Cref{sec: wpb_gauss}. The underlying proof technique is general enough to deal with (possibly unbounded) convex smooth loss functions. More details are gathered in \Cref{th: main_gaussian_smooth,cor: unbounded_smooth}.

\textit{A WPB bound with data-dependent prior.}
As we aim to intricate optimisation guarantees with generalisation bounds, we have to overcome the Bayesian paradigm of data-free priors which sets the prior distribution as a comparison point. Here, it is necessary to compare the candidate posterior with the optimisation goal. To do so, we elaborate in \Cref{sec: data_dep_priors} on the idea of \citet{dziugaite2018data} who exploit differential privacy to obtain PAC-Bayesian bounds allowing to take data-dependent priors. We show that it is possible to maintain the asymptotic convergence rate of \Cref{cor: unbounded_lpz} when taking as 'prior' a Gibbs posterior.
We introduce the following theorem holding again when $\mathcal{H}=\mathbb{R}^d$. For any gaussian prior $P$ living in $C_{\alpha,\beta,M}$, with probability at least $1-\delta$, for any posterior distribution $Q\in C_{\alpha,\beta,M}$, we have the following asymptotic convergence rate
\begin{align*}
|\Delta_S(Q)|  \leq\Tilde{\mathcal{O}}\left( \sqrt{2K\frac{d}{m}\left(1 + W_1(Q,P_{-\frac{\lambda}{2K}R_S})\right)+(1+K^2\log(m))\frac{\log\left( \frac{m}{\delta} \right)}{m}}   \right).
\end{align*}
We also study non-asymptotic regimes in \Cref{th: data_dep}. While \citet{dziugaite2018data} exploited differential privacy results for the Gibbs posterior when the loss is bounded, we successfully extended these results to (possibly unbounded) uniformly Lipschitz losses. This is not specific to the WPB framework and may be of independent interest.

\textit{PAC-Bayes provides generalisation guarantees for the Bures-Wasserstein SGD.}
While working on WPB theory, we notice a shift from classical assumptions due to the KL divergence. Indeed, statistical assumptions (such as subgaussiannity, bounded variances) are transformed into geometric assumptions such as Lipschitzness and convex smoothness when Wasserstein distances are involved. We exploit in \Cref{sec: gene_sgd} WPB theory to provide generalisation guarantees for the Bures-Wasserstein SGD (recalled in \Cref{alg: sgd}) which approximates the best Gaussian surrogate $\hat{Q}$ of $Q^*:= P_{-\frac{\lambda}{2K}R_S}$ (in the sense of the KL divergence, see \Cref{sec: gene_sgd} for more details).
More precisely, we show that the KL divergence and Wasserstein distances are linked within the WPB framework: the (KL-based) PAC-Bayesian learning objective of \citet{catoni2007pac}, which outputs the Gibbs posterior $Q^*$, can be approximated by $\hat{Q}_N$, the output of the Bures-Wasserstein SGD after $N$ optimisation steps, which is provably close from $\hat{Q}$ with respect to the $2$-Wasserstein distance (see \Cref{th: lambert}).
Within the WPB framework, this link is translated in \Cref{th: main_sgd} as a generalisation bound ensuring that asymptotically, the minima reached by the Bures-Wasserstein SGD has a strong generalisation ability.
\medskip

Concretely, for $N$ large enough, for uniformly $K$-lipschitz, convex, smooth loss functions we have the following asymptotic guarantee with probability $1-\delta$:
\begin{align*}
|\Delta_S(\hat{Q}_N)|  \leq\Tilde{\mathcal{O}}\left( \sqrt{2K\frac{d}{m}\left(1 + W_1(\hat{Q},Q^*)\right)+ (1+K^2\log(m)) \frac{\log\left( \frac{m}{\delta} \right)}{m}} \right).
\end{align*}
Thus, the WPB framework is enough to provide an explicit convergence rate for the generalisation gap avoiding the comparison to an arbitrary prior. Instead, this bound shows that a (long enough) run of the Bures-Wasserstein SGD with enough data (or a Lipschitz constant small enough) leads to a minimiser with a high generalisation ability. Furthermore, \Cref{th: main_sgd} is a reformulation of \eqref{eq: complete_bound_sgd} which is, to our knowledge, the first PAC-Bayesian bound of the form \eqref{eq: wanted_pattern} with $D=\sqrt{d W_2}$ and $$\varepsilon= \mathcal{O}(\sqrt{dW_1(\hat{Q},Q^*)}).$$
This provides elements of answer to the question listed in \Cref{sec: intro_optim} and concludes this work.

\begin{figure}[ht]
\label[figure]{fig: overview}
\centering
\includegraphics[scale=0.5]{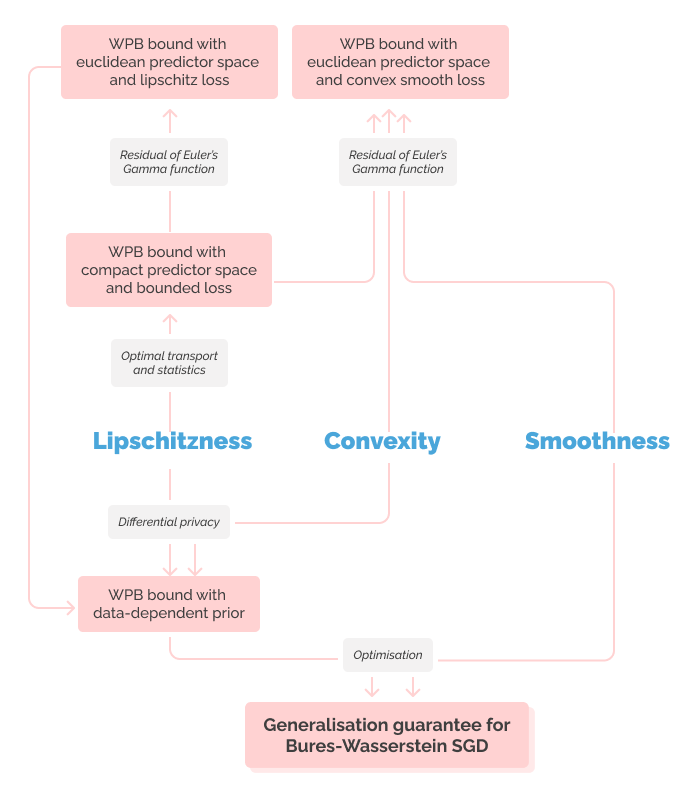}
\caption{An overwiew of the assumptions required to obtain the main results. Assumptions are stated in blue, main results are in pink boxes and the proof technique exploited to obtain such results are within grey boxes.}
\end{figure}

\textit{Discussion about the assumptions}
For the sake of clarity, we provide in \Cref{fig: overview} the topography of our main results. We focus on the assumptions required to state each of the results and doing so, we aim to give to the reader a broader vision of when can these bounds be applied. We stress that the Lipschitzness assumption is at the core of all results, except \Cref{th: main_gaussian_smooth,cor: unbounded_smooth}.
Convexity is required to use differential privacy and to obtain \Cref{th: data_dep}. Finally, we note that while the results of \citet{lambert2022variational} are usable with only smoothness and convexity, we must add the uniform Lipschitz assumption to obtain \Cref{th: main_sgd}. The question of whether all these assumptions are minimal to perform WPB remains open.

\section{PAC-Bayesian bounds for compact predictor spaces}
\label{sec: compact_space}

Here we establish WPB bounds for bounded losses when the predictor space is a compact of $\mathbb{R}^d$. To intricate the $1$-Wasserstein distance within the PAC-Bayes proof, we design a surrogate of the change of measure inequality \citep{donsker1975asymptotic} by exploiting the uniform Lipschitz assumption on the loss. To do so we need to exploit the notion of \emph{covering number} recalled below as well as Kantorovich duality \citep[Theorem 5.10]{villani2009optimal}.
This notion of duality holds for any cost function (in an optimal transport framework) contrary to the Kantorovich-Rubinstein duality exploited by \citet{amit2022ipm} which only holds when the cost function is a distance. This result is recalled in \Cref{sec: back_compact}.
\begin{definition}[Covering number]
Let $\mathcal{H}\subseteq \mathbb{R}^d$. An $\varepsilon$-covering of $\mathcal{H}$ is a subset $C$ of $\mathcal{H}$ such that $\mathcal{H} \subseteq \cup_{x\in C} \bar{\mathcal{B}}(x,\varepsilon)$. The $\varepsilon$-covering number of $\mathcal{H}$ is defined as
$$N(\mathcal{H},\varepsilon):= \min\{n\geq 1 \mid \exists \text{ an $\varepsilon$-covering of $\mathcal{H}$ of size $n$} \}.$$
\end{definition}
We also define the $\varepsilon,1$-Wasserstein to be $W_{\varepsilon}(Q,P)= \varepsilon + W_1(Q,P)$. This cost function is essential to the analysis.
We now state the main results of this section. Additional background is gathered in \Cref{sec: back_compact}.

\subsection{A Catoni-type bound}

We propose here a WPB bound analogous to a relaxation of \citet[Theorem 1.2.6]{catoni2007pac} stated for instance in \citet[Theorem 4.1]{alquier2016variational}.

\begin{theorem}
\label{th: compact_catoni}
For any $\varepsilon,\delta>0$, assume that $\ell\in [0,1]$ is uniformly $K$-Lipschitz and that $\mathcal{H}$ is a compact of $\mathbb{R}^d$ bounded by $R>0$. Let $P\in \mathcal{P}_1(\mathcal{H})$ be a (data-free) prior distribution and assume we choose a parameter $\lambda$ such that
\[ 0< \lambda \leq  \frac{1}{K}\sqrt{\frac{2m}{2d\log(1+\frac{2R}{\varepsilon})+\log(\frac{2}{\delta})}}:= \lambda_{max}. \]
Then, with probability $1-\delta$ , for any posterior distribution $Q\in\mathcal{P}_1(K)$,
\[ \Delta_S(Q) \leq 4K\varepsilon + \frac{W_1(Q,P)+2\varepsilon +\log(2/\delta)}{\lambda} + \frac{\lambda}{2m}.   \]
\end{theorem}
Note that we assumed the loss to be bounded, although this can be relaxed to subgaussiannity at no cost.
In \Cref{th: compact_catoni}, the range of $\lambda$ is restricted and the loss required to be uniformly Lipschitz. Such restrictions do not exist in \citet[Theorem 4.1]{alquier2016variational} which recovers a similar result with a KL divergence coming from the change of measure inequality \citep{donsker1975asymptotic}. In WPB this is required to have a control on $\Delta_S$ which is exploited in Kantorovich duality (\Cref{th: kanto_dual}).
Furthermore, assuming Lipschitzness on a compact space is not restrictive as it covers, \emph{e.g.}, all $\mathcal{C}^1$ functions.
Note that the smaller the Lipschitz constant $K$ is, the larger $\lambda_{max}$.
This is not surprising as, from an optimisation point of view, $\lambda$ acts as a learning rate which determines the influence of data with respect to the regulariser $W_1(Q,P)$.
A small $K$ says that huge variations between data have a small influence on the loss value, then we can give more influence to the training set without deteriorating much the generalisation ability of the posterior.
This bound also says that it is legitimate to consider a WPB learning objective analogous to the one derived from \citet[Theorem 4.1]{alquier2016variational} (which yields Gibbs posteriors):
$$\operatorname{argmin}_{Q\in\mathcal{P}_1(\mathbb{\mathcal{H}})}\frac{W_1(Q,P)}{\lambda} + \frac{\lambda}{2m}.$$
\Cref{th: compact_catoni}'s proof is stated below and mixes up several arguments from optimal transport with PAC-Bayes learning through covering numbers.
\begin{proof}[Proof of \Cref{th: compact_catoni}]

\textit{Step 1: define a good data-dependent function.} We define, for any sample $S$ and predictor $h\in \mathcal{H}$,
\[ f_S(h) = \lambda \Delta_S(h). \]
This function satisfies the following lemma:
\begin{lemma}
\label[lemma]{l: quasi_lpz_func}
Let $\varepsilon>0$ assume that $0<\lambda\leq \frac{1}{K}\sqrt{\frac{2m}{\log\left(\frac{N(\mathcal{H},\varepsilon)^2}{\delta}\right)}} $. We have, with probability $1-\delta$ for all $h,h'\in\mathcal{H}$, for any $P$:

\[f_S(h)-f_S(h') \leq 2(1+2\lambda K)\varepsilon + ||h-h'||.  \]

\end{lemma}
\begin{proof}[Proof of \Cref{l: quasi_lpz_func}]
We rename here $N:= N(\mathcal{H},\varepsilon)$. There exists an $\varepsilon$-covering $C:=\{h_1,...,h_N\}$ of $\mathcal{H}$ of size $N$.
Then for any $h,h'\in C^2$, we have:
\begin{align*}
f_S(h)-f_S(h')  & = \frac{\lambda}{m}\sum_{i=1}^m \mathbb{E}[\ell(h,z)-\ell(h'z)] - \left(\ell(h,z_i)-\ell(h',z_i))\right).
\end{align*}
We know that for any $h,h',z$, $|\ell(h,z)-\ell(h',z)| \leq \lambda K||h-h'||$. Then, applying Hoeffding's inequality for all pairs $h,h'\in C^2$ and performing an union bound gives that with probability at least $1-\delta$, for all pairs $(h,h')\in C^2$ :
\begin{align*}
f_S(h)-f_S(h')  \leq \sqrt{\frac{\log\left(\frac{N^2}{\delta}\right)}{2m}} \lambda K ||h-h'||.
\end{align*}
So for any $h,h'\in \mathcal{H}^2$ there exists $h_0,h'_0\in C^2$ such that $||h-h_0||\leq \varepsilon$ and $||h-h_0||\leq \varepsilon$. Thus, we have
\begin{align*}
f_S(h)-f_S(h')  & = f_S(h)-f_S(h_0) + f_S(h_0)-f_S(h'_0)+ f_S(h'_0)-f_S(h') \\
& \leq 2\lambda K \left(||h-h_0|| + ||h'-h'_0||\right) + \sqrt{\frac{\log\left(\frac{N^2}{\delta}\right)}{2m}} \lambda K ||h_0-h_0'|| \\
& \leq 4\lambda K \varepsilon + \sqrt{\frac{\log\left(\frac{N^2}{\delta}\right)}{2m}} \lambda K ||h_0-h'_0||.
\end{align*}
By the triangle inequality, $||h_0-h'_0||\leq ||h-h'|| + 2\varepsilon$ so we finally have with probability at least $1-\delta$, for any $h,h'\in K^2$:
\[  f_S(h)-f_S(h') \leq 4\lambda K \varepsilon + \sqrt{\frac{\log\left(\frac{N^2}{\delta}\right)}{2m}} \lambda K \left( 2\varepsilon+ ||h-h'||\right).\]
Using $\lambda \leq \frac{1}{K}\sqrt{\frac{2m}{\log\left(\frac{N^2}{\delta}\right)}}$ and upper bounding concludes the proof.
\end{proof}

\textit{Step 2: A probabilistic change of measure inequality for $f_S$.}
We do not have for the Wasserstein distance such a powerful tool than the change of measure inequality. However, we can generate a probabilistic surrogate on $\mathcal{P}_1(\mathcal{H})$ valid for the function $f_S$.

\begin{lemma}
\label[lemma]{l: change_meas}
For any $\epsilon>0$, any $\delta>0$, any $$0<\lambda\leq \frac{1}{K}\sqrt{\frac{2m}{\log\left(\frac{N(\mathcal{H},\varepsilon)^2}{\delta}\right)}},$$
we have with probability $1-\delta$ over the sample $S$, for any $P\in\mathcal{P}_1(K)$
\[ \left(\sup_{Q\in \mathcal{P}_1(K)} \mathbb{E}_{h\sim Q}[ f_S(h)] - (2(1+\lambda K)\varepsilon - W_1(Q,P) \right) \leq \mathbb{E}_{h\sim P}[ f_S(h)].     \]
\end{lemma}

\begin{proof}[Proof of \Cref{l: change_meas}]
Firstly, we introduce the cost function $c_{\varepsilon}(x,y)= \varepsilon + ||x-y||$.
From this we notice that we can rewrite the $\varepsilon,1$- Wasserstein distance:
\[ W_{\varepsilon}(Q,P)= \inf_{\pi \in \Pi(Q,P)} \int_{\mathcal{H}^2} c_{\varepsilon}(x,y)\mathrm{d}\pi(x,y).  \]
Remark that because $W_1$ is a distance, then $W_\varepsilon$ is symmetric. Furthermore, if we fix $\mathcal{X}=\mathcal{Y}=\mathcal{H}$ and we notice that $c_{\varepsilon}\geq 0$, then the condition for Kantorovich duality is satisfied. Thus, we apply \Cref{th: kanto_dual} as follows: for all $Q,P\in \mathcal{P}_1(\mathcal{H})$:
\begin{align*}
W_\varepsilon(Q,P)= W_\varepsilon(P,Q) & =  \min _{\pi \in \Pi(P, Q)}  \int_{K^2} c_{\varepsilon}(h_1, h_2) d \pi(h_1, h_2)  \\
&=\sup_{\substack{(\psi, \phi) \in L^1(Q) \times L^1(P)\\ \psi-\phi \leq c_{\varepsilon}}}\left[\int_{K} \psi(h) \mathrm{d}Q(h)- \int_{K} \phi(h) \mathrm{d}P(h)\right] \\
& = \sup_{\substack{(\psi, \phi) \in L^1(Q) \times L^1(P)\\ \psi-\phi \leq c_{\varepsilon}}}\left[\mathbb{E}_{h\sim Q}[ \psi(h)]- \mathbb{E}_{h\sim P}[ \phi(h)]\right].
\end{align*}
A crucial point is that for a well-chosen $\lambda$ with high probability, the pair $(f_S,f_S)$ satisfies the condition stated under the last supremum. It is formalised in the following lemma.
\begin{lemma}
\label{lem:kk}
For any $\varepsilon>0$ any $\delta>0$, any $0<\lambda\leq \frac{1}{K}\sqrt{\frac{2m}{\log\left(\frac{N(\mathcal{H},\varepsilon)^2}{\delta}\right)}}$ , we have with probability at least $1-\delta$ over the sample $S$ that, for all measures $Q,P\in\mathcal{P}_1(\mathcal{H})^2$:
\begin{itemize}
  \item $f_S\in L_1(Q),L_1(P)$,
  \item for all $h,h' \in \mathcal{H}^2, f_S(h)-f_S(h') \leq c_{\varepsilon'}(h,h')$ with $\varepsilon':= 2(1+2\lambda K) \varepsilon$.
\end{itemize}
Thus, Kantorovich duality (\Cref{th: kanto_dual}) gives:
\[ \left(\sup_{Q\in \mathcal{P}_1(\mathcal{H})} \mathbb{E}_{h\sim Q}[ f_S(h)] -  W_{\varepsilon'}(Q,P) \right)\leq \mathbb{E}_{h\sim P}[ f_S(h)],    \]
and using $W_{\varepsilon'} = \varepsilon' + W_1$ and the definition of $\varepsilon'$ concludes the proof.
\end{lemma}
\begin{proof}[Proof of \Cref{lem:kk}]
Because the space of predictors $\mathcal{H}$ is compact and that for any $z\in\mathcal{Z}$, the loss function $\ell(.,z)$ is $K$-Lipschitz on $\mathcal{H}$, then both the generalisation and empirical risk are continuous on $\mathcal{H}$. Thus $|f_S|$ is also continuous and, by compacity, reaches its maximum $M_S$ on $\mathcal{H}$. Thus for any probability $P$ on $\mathcal{H}, \mathbb{E}_{h\sim P}[|f_S(h)|] \leq M_S < +\infty$ almost surely. This proves the first statement.
We notice that the second statement, given the choice of $\lambda$, is the exact conclusion  of \Cref{l: quasi_lpz_func} with probability at least $1-\delta$.
So with probability at least $1-\delta$, Kantorovich duality gives us that for any $P,Q$ with $\varepsilon'= 2(1+ \lambda K) \varepsilon$,
\begin{align*}
\mathbb{E}_{h\sim Q}[ f_S(h)] - \mathbb{E}_{h\sim P}[f_S(h)] \leq W_{\varepsilon'}(Q,P).
\end{align*}
Re-organising the terms and taking the supremum over $Q$ concludes the proof.
\end{proof}
This concludes the proof of \Cref{l: change_meas}.
\end{proof}

\textit{Step 3: The PAC-Bayes route of proof for the 1-Wasserstein distance.}

We start by exploiting \Cref{l: change_meas}: for any prior $P\in\mathcal{P}_1(K)$, for $$0<\lambda \leq  \frac{1}{K}\sqrt{\frac{2m}{\log\left(\frac{2N(K,\varepsilon)^2}{\delta}\right)}},$$
with probability at least $1-\delta/2$ we have
\[ \left(\sup_{Q\in \mathcal{P}_1(K)} \mathbb{E}_{h\sim Q}[ f_S(h)] - (2(1+2\lambda K)\varepsilon - W_1(Q,P) \right) \leq \mathbb{E}_{h\sim P}[ f_S(h)]. \]
We then notice that by Jensen's inequality,  $$\mathbb{E}_{h\sim P}[ f_S(h)] \leq \log\left(\mathbb{E}_{h\sim P}[ \exp(f_S(h))]    \right).$$
Then, by Markov's inequality we have with probability $1-\delta/2$
\[ \mathbb{E}_{h\sim P}[ f_S(h)] \leq \log\left(\frac{2}{\delta}\right) + \log\left(\mathbb{E}_S\mathbb{E}_{h\sim P}\left[ \exp(f_S(h))   \right]\right).  \]
By Fubini and Hoeffding lemma applied $m$ times on the iid sample $S$, we have
\[ \mathbb{E}_S\mathbb{E}_{h\sim P}\left[ \exp(f_S(h))\right] =  \mathbb{E}_{h\sim P}\mathbb{E}_S\left[ \exp(f_S(h))\right] \leq \frac{\lambda^2}{2m}. \]
Taking an union bound gives us with probability $1-\delta$, for any posterior $Q$:
\[ \mathbb{E}_{h\sim Q}[R(h)] \leq \mathbb{E}_{h\sim Q}[R_m(h)] +   4K\varepsilon + \frac{W_1(Q,P)+2\varepsilon +\log(2/\delta)}{\lambda} + \frac{\lambda}{2m}.   \]
Finally, we know that $\mathcal{H}$ is bounded by $R$ so by \Cref{prop: covering} we have $$N^2 = N(\bar{\mathcal{B}}(0,R), \varepsilon)^2 \leq \left(1+ 2mR \right)^{2d}.$$
Thus, we can take $\lambda$ equal to $$\frac{1}{K}\sqrt{\frac{2m}{2d\log(1+\frac{2R}{\varepsilon})+\log(\frac{2}{\delta})}}.$$
This concludes the proof.
\end{proof}

\subsection{A McAllester-type bound}
We now move on to a McAllester-type bound, which can be tighter than \Cref{th: compact_catoni} for large values of the $1$-Wasserstein.

\begin{theorem}
\label{th: compact_mcall}
For any $\delta>0$, assume that $\ell\in[0,1]$ is uniformly $K$-Lipschitz and that $\mathcal{H}$ is a compact of $\mathbb{R}^d$. Let $P\in \mathcal{P}_1(\mathcal{H})$ a (data-free) prior distribution.
Then, with probability $1-\delta$ , for any posterior distribution $Q\in\mathcal{P}_1(\mathcal{H})$:
\[ |\Delta_S(Q)| \leq \sqrt{2K(2K+1)\frac{2d\log\left(3\frac{1 +2Rm }{\delta}\right)}{m} \left(W_1(Q,P)+\varepsilon_m  \right) + \frac{\log\left( \frac{3m}{\delta} \right)}{m}  }, \]
with $\varepsilon_m = \frac{4}{\log(\frac{3}{\delta})} \left( 2 + \sqrt{\frac{\log\left(\frac{3}{\delta}\right) + 2d\log(1+2Rm)}{2m}}  \right) = \mathcal{O}\left(1 + \sqrt{\nicefrac{d\log(Rm)}{m}}\right)$.
\end{theorem}
We deteriorate the bound of \cite{amit2022ipm} by transforming a convergence rate of $$\sqrt{\frac{W_1(Q,P)}{m}}$$ for finite predictor classes onto a  $\sqrt{\left(Kd W_1(Q,P)+1\right)\frac{\log(m)}{m}}$ for compact classes. This deteriorated rate is the price to pay to consider a general WPB bound for an uncountable number of predictors. However, notice that the dimension dependency can be attenuated through the Lipschitz constant, with the limit rate of $$\mathcal{O}\left(\sqrt{\nicefrac{\log\left( \frac{m}{\delta} \right)}{m}}\right)$$ which is dimension-free and is a consequence of the statistical component of PAC-Bayes learning. Furthermore, note that this proof technique allows us to recover the rate of \citet{amit2022ipm} rate when considering finite classes.
The proof of \Cref{th: compact_mcall} involves similar arguments to the one of \Cref{th: compact_catoni}, therefore we defer it to \Cref{sec: proof_compact_mcall}.

\section{PAC-Bayesian bounds for Gaussian distributions}
\label{sec: wpb_gauss}

In this section we develop McAllester-type WPB bounds on an Euclidean predictor space. Indeed, in PAC-Bayes learning, considering this predictor space is common as PAC-Bayesian objective often focuses on Gaussian priors and posteriors (see, \emph{e.g.}, \citealp{dziugaite2017computing,amit2018meta,haddouche2022online}).
Those bounds build up on \Cref{th: compact_mcall} and the overall conclusion is the following: when considering functions with interesting geometric properties (\emph{i.e.}, Lipschitzness or smoothness) on $\mathbb{R}^d$, WPB bounds hold for Gaussian priors and posteriors over $\mathcal{H}= \mathbb{R}^d$ at the cost of negligible extra terms (\Cref{th: main_gaussian_lpz,th: main_gaussian_smooth}).
More importantly, we show that in this setup, the assumption of a bounded loss is not required anymore to perform WPB: only boundedness on a compact is needed. Thus, we propose WPB bounds for unbounded losses (\Cref{cor: unbounded_lpz,cor: unbounded_smooth}).

\textit{Two sets of assumption.} Previously, we assumed two assumptions on the losses: uniform Lipschitzness (\Cref{def: unif_lpz}) and boundedness (in $[0,1]$) on a compact of $\mathbb{R}^d$. We provide below to novel sets of hypotheses which encapsulates previous assumptions while allowing the loss to be unbounded on all $\mathbb{R}^d$.
\begin{itemize}
  \item \textbf{(A1)} $\ell$ is uniformly $K$-Lipschitz over  $\mathcal{H}$, and $\sup_{z\in\mathcal{Z}} || \ell(0,z)|| =D< +\infty.$
  \item \textbf{(A2)} For any $z\in\mathcal{Z}$, $\ell(.,z)$ is continuously differentiable over $\mathcal{H}$, $\ell(.,z)$ is also a convex $L$- smooth (\emph{i.e}, its gradient is $L$-Lipschitz) and $\sup_{z\in\mathcal{Z}} ||\nabla_h \ell(0,z)|| =D< +\infty$.
\end{itemize}
\begin{example}
Recall that $\mathcal{H}=\mathbb{R}^d$ and let $\phi:\mathcal{H}\rightarrow \mathbb{R}^d$.
Also, let $\psi :\mathcal{Z}\rightarrow \mathbb{R}^d$ such that $\psi(\mathcal{Z})$ is bounded by $C_\phi >0$. We assume that both $\phi,\psi$ are continuously differentiable and that $\nabla\phi$ is $G$-Lipschitz.
Note that the $||\phi||$ is possibly unbounded on $\mathcal{H}$.
Then \textbf{(A2)} holds for the loss function $\ell(h,z)= ||\phi(h)-\psi(z)||^2$
Indeed, $\nabla_h \ell(h,z)= 2(\nabla\phi (h) -\psi(z))$ so on any compact $\mathcal{K}$ bounded by $R$, $\nabla_h \ell$ is uniformly $2$-Lipschitz. Also $\sup_{z\in\mathcal{Z}} ||\nabla_h \ell(0,z)|| \leq 2C$.
Note that on $\mathbb{R}^d$, $\ell(.,z)$ is not necessarily Lipschitz for any $z$ (take the case $\phi= Id_{\mathbb{R}^d}$) so \textbf{(A1)} is not satisfied.
\end{example}

\textit{A brief summary of the proof technique.} To extend \Cref{th: compact_mcall} to the case $\mathcal{H}=\mathbb{R}^d$, we use the push-forward distribution $\mathcal{P}_R \# P$ where $P\in C_{\alpha,\beta,M}$ for fixed $\alpha,\beta,M$ (notation defined in \Cref{sec: framework}).
The interest of this is to use \Cref{th: compact_mcall} by considering projections of the Gaussian prior and posterior. When considering Gaussian distributions, the gap between projected distributions and original ones is explicitly controlled.
More precisely, for any $R>0$ large enough, for any $P\in C_{\alpha,\beta, M}$, $W_1(P, \mathcal{P}_R\# P)$ is upper bounded. This is the conclusion of an important technical lemma (\Cref{l: gaussian_tail}), stated with additional background in \Cref{sec: background_gaussian}.
We state below new WPB results with Gaussian distributions for Lipschitz functions in \Cref{sec: main_sec_gaussian_lpz} and for smooth functions in \Cref{sec: main_sec_gaussian_smooth}.

\subsection{PAC-Bayesian bounds for Lipschitz losses}
\label{sec: main_sec_gaussian_lpz}

This section focuses on the case of Lipschitz losses. We show that when the loss is uniformly Lipschitz, it is possible to maintain the tightness of \Cref{th: compact_mcall} on all $\mathbb{R}^d$ when the loss remains bounded. We also show that it is also possible to obtain a WPB bound when the loss function satisfies \textbf{(A1)} (\emph{i.e.} with an additional boundedness assumption on $\sup_{z} \ell(0,z)$), while remaining unbounded (\Cref{cor: unbounded_lpz}).

\begin{theorem}
\label{th: main_gaussian_lpz}
Assume that $d\geq 3$, $\mathcal{H}= \mathbb{R}^d$ and that the loss is uniformly $K$-Lipschitz and lies in $[0,1]$ over $\mathcal{H}$ . For any $\delta>0, 0\leq \alpha\leq \beta, M\geq 0$, let $P\in C_{\alpha,\beta,M}$ a (data-free) prior distribution.
Then, with probability $1-\delta$ , for any posterior distribution $Q\in C_{\alpha,\beta,M}$:
\begin{align*}
|\Delta_S(Q)|  \leq 2\frac{\beta{\sqrt{\beta}}}{m} + \sqrt{2K(2K+1)\frac{2d\log\left(3\frac{1 +2Rm }{\delta}\right)}{m} \left(W_1(Q,P)+\alpha_m \right) + \frac{\log\left( \frac{3m}{\delta} \right)}{m}  },
\end{align*}
with $R=\mathcal{O}(\max \sqrt{d\log(d)},\sqrt{\log(m)})$ and $\alpha_m= 2(M+1)\frac{\beta\sqrt{\beta}}{m} + \varepsilon_m= \mathcal{O}\left(1 + \sqrt{\nicefrac{d\log(Rm)}{m}}\right)$ with $\varepsilon_m$ defined in \Cref{th: compact_mcall}.
\end{theorem}
\Cref{th: main_gaussian_lpz} shows that, at the cost of additional residual terms, it is possible to maintain the convergence rate of \Cref{th: compact_mcall} when considering Gaussian prior and posterior within the compact $C_{\alpha,\beta,M}$. The influence of $\alpha,\beta,\gamma$ appear in the explicit value of $R$ described as it is always taken in this work as the smallest value satisfying the assumption \texttt{Rad} described in \Cref{sec: background_gaussian}.
As in \Cref{th: compact_mcall}, the idea that a small Lipschitz constant tightens the bound is still conveyed here and is of great importance for \Cref{cor: unbounded_lpz} which provides a WPB bound for unbounded losses with higher dimension dependency when few data is available.
\begin{proof}[Proof of \Cref{th: main_gaussian_lpz}]
We take a specific radius $R$ which is the smallest value satisfying \texttt{Rad}.
The proof starts with a straightforward application of \Cref{th: compact_mcall} on  the compact $\mathcal{B}(0,R)$, with the prior $\mathcal{P}_R\# P$, and with high probability, for any posterior $\mathcal{P}_R\# Q$ with $Q\in C_{\alpha,\beta,M}$:
\begin{align*}
|\Delta_S(\mathcal{P}_R\# Q)|  \leq  \sqrt{2K(2K+1)\frac{2d\log\left(3\frac{1 +2Rm }{\delta}\right)}{m} \left(W_1(\mathcal{P}_R\# Q,\mathcal{P}_R\# P) + \varepsilon_m \right) + \frac{\log\left( \frac{3m}{\delta} \right)}{m} }.
\end{align*}
From this we control the left hand-side term as follows:
\begin{align*}
|\Delta_S(Q)| &\leq |\Delta_S(\mathcal{P}_R\# Q)| + |\Delta_S(Q) - \Delta_S(\mathcal{P}_R\# Q) |.
\intertext{And we also have}
|\Delta_S(Q) - \Delta_S(\mathcal{P}_R\# Q) | & \leq \mathbb{E}_{h\sim Q}\left[|\Delta_S(h) - \Delta_S(\mathcal{P}_R(h))|\right] \\
& = \mathbb{E}_{h\sim Q}\left[|\Delta_S(h) - \Delta_S(\mathcal{P}_R(h))| \mathds{1}(||h||>R)\right] \\
&\leq 2Q(||h||>R) \leq 2 \frac{\beta\sqrt{2\beta}}{m},
\end{align*}
the last line holding thanks to \Cref{l: gaussian_tail} and because $\Delta_S\in[-1,1]$.
Also we have by the triangle inequality:
\[W_1(\mathcal{P}_R\# Q,\mathcal{P}_R\# P) \leq W_1(Q, \mathcal{P}_R\# Q) + W_1(Q,P)+ W_1(P, \mathcal{P}_R\# P). \]
Because both $Q,P\in C_{\alpha,\beta,M}$, using again \Cref{l: gaussian_tail} gives:
\[W_1(\mathcal{P}_R\# Q,\mathcal{P}_R\# P) \leq  W_1(Q,P)+2(M+1)\frac{\beta\sqrt{2\beta}}{m}. \]
We then have:
\begin{align*}
|\Delta_S(Q)| \leq 2 \frac{\beta\sqrt{2\beta}}{m} + \sqrt{2K(2K+1)\frac{2d\log\left(3\frac{1 +2Rm }{\delta}\right)}{m} \left(W_1(Q,P)+ \alpha_m \right) + \frac{\log\left( \frac{3m}{\delta} \right)}{m} },
\end{align*}
with $\alpha_m= 2(M+1)\frac{\beta\sqrt{\beta}}{m} + \varepsilon_m= \mathcal{O}(1)$. This concludes the proof.
\end{proof}

\textit{A corollary for unbounded losses.} We provably extend \Cref{th: main_gaussian_lpz} to the case of unbounded Lipschitz losses.

\begin{corollary}
\label[corollary]{cor: unbounded_lpz}
Assume that $d\geq 3$, $\mathcal{H}= \mathbb{R}^d$ and that the (unbounded) loss satisfies \textbf{(A1)}. For any $\delta>0, 0\leq \alpha\leq \beta, M\geq 0$, let $P\in C_{\alpha,\beta,M}$ a (data-free) prior distribution.
Then, with probability $1-\delta$, for any posterior distribution $Q\in C_{\alpha,\beta,M}$, the three following bounds holds.

\noindent \textbf{Low-data regime} $(d\geq m)$
\begin{align*}
|\Delta_S(Q)|  \leq\Tilde{\mathcal{O}}\left( \sqrt{2K\frac{d^{\nicefrac{3}{2}}}{m}\left(\sqrt{\frac{d}{m}}+ W_1(Q,P)\right)+ (1+K^2d)\frac{\log\left( \frac{m}{\delta} \right)}{m}}   \right).
\end{align*}
\noindent \textbf{Transitory regime} ($m>d,\; d\log(d)\geq \log(m)$)

\begin{align*}
|\Delta_S(Q)|  \leq\Tilde{\mathcal{O}}\left( \sqrt{2K\frac{d^{\nicefrac{3}{2}}}{m}\left(1+ W_1(Q,P)\right) + (1+K^2d)\frac{\log\left( \frac{m}{\delta} \right)}{m}}   \right).
\end{align*}

\noindent \textbf{Asymptotic regime} ($d\log(d)< \log(m)$)
\begin{align*}
|\Delta_S(Q)|  \leq\Tilde{\mathcal{O}}\left( \sqrt{2K\frac{d}{m}\left(1 + W_1(Q,P)\right)+ (1+K^2\log(m))\frac{\log\left( \frac{m}{\delta} \right)}{m}}   \right).
\end{align*}
In all these formulas, $\Tilde{\mathcal{O}}$ hides a polynomial dependency in $(\log(d),\log(m))$.
For an explicit formulation of the bounds, we refer to \eqref{eq: complete_lpz_unbounded}.
\end{corollary}
The message here is that in Wasserstein PAC-Bayes, the bounded loss assumption is not as important as in classical PAC-Bayes using KL divergence. Indeed, the geometric constraints of WPB forced us to consider compact classes of Gaussian distribution and Lipschitz losses. Having such geometric assumptions on the distribution space and the loss is enough to exploit the properties of the $1$-Wasserstein distance and to circumvent the boundedness assumption.
To avoid boundedness, we transformed the limit rate $$\mathcal{O}\left(\sqrt{\nicefrac{\log\left( \frac{m}{\delta} \right)}{m}}\right)$$ of \Cref{th: compact_mcall} into $$\mathcal{O}\left(\sqrt{(1+K^2  d)\nicefrac{\log\left( \frac{m}{\delta} \right)}{m}}\right)$$ for non-asymptotic regimes
and $$\mathcal{O}\left(\sqrt{(1+K^2\log(m))\nicefrac{\log\left( \frac{m}{\delta} \right)}{m}}\right)$$ for the asymptotic one.
Thus, even when few data is available, a well constrained (unbounded) Lipschitz loss is able to control the impact of the dimension.
Note that, in the small data regime, we have the highest dimension dependency. Note also that the dimensionality of the learning problem is controlled by the Lipschitz constant with the limit rate of $$\mathcal{O}\left(\sqrt{\nicefrac{\log\left( \frac{m}{\delta} \right)}{m}}\right)$$ which is dimension-free and is a consequence of the statistical component of PAC-Bayes learning.
To the best of our knowledge, our work is the first to exploit geometric properties of the loss to propose PAC-Bayes bounds for unbounded and heavy-tailed losses with explicit convergence rates. Indeed, the existing literature on unbounded losses exploits either general divergence properties \citep{alquier2017simpler,picard2022divergence}, functional properties for heavy-tailed distribution \citep{holland2019}, uniform boundedness assumption on the loss over the data space \citep{haddouche2021pac} or concentration inequalities \citep{kuzborskij2019efron,rivasplata2020pac,haddouche2022pac,jang2023tight}.

\begin{proof}[Proof of \Cref{cor: unbounded_lpz}]
Firstly, we start from  \Cref{th: compact_mcall} which gives, with probability at least $1-\delta$:
\begin{equation}
\label{eq: tight_mcall}
|\Delta_S(Q)| \leq \sqrt{ 2K(2K+1) \frac{\log(\frac{3}{\delta}) + 2d\log\left(1 +2Rm \right)}{m} \left(W_1(Q,P)+\varepsilon_m \right) + \frac{\log\left( \frac{3m}{\delta} \right)}{m} }.
\end{equation}
This last bound holds for any uniformly Lipschitz function taking value on $[0,1]$ on a compact predictor space bounded by a certain $R$.
Let $P\in C_{\alpha,\beta,M}$.
We now assume \textbf{(A1)} and consider $R$ to be the smallest value satisfying \texttt{Rad}.
Let $\ell'= \ell/(D+2KR)$. We note $D_R= D + 2KR$, then on the ball $\mathcal{B}(0,R)$, $\ell'$ takes value in $[0,1]$ (because the compact is bounded by $R$ and the loss is $K$-Lipschitz) and is $K/D_R$-Lipschitz.
Applying \Cref{eq: tight_mcall} with $\ell'$ on $\mathcal{B}(0,R)$ and multiplying by $D_R$ gives, with high probability, for any $Q\in C_{\alpha,\beta,M}$:
\begin{multline*}
|\Delta_S(\mathcal{P}_R\# Q)| \\ \leq D_R\sqrt{ 2 \frac{K}{D_R} (2 \frac{K}{D_R} +1) \frac{\log(\frac{1}{\delta}) + 2d\log\left(1 +2Rm \right)}{m} \left(W_1(\mathcal{P}_R\# Q,\mathcal{P}_R\# P)+\varepsilon_m \right) + \frac{\log\left( \frac{m}{\delta} \right)}{m} } \\
=\sqrt{ 2 K(2 K+D_R) \frac{\log(\frac{1}{\delta}) + 2d\log\left(1 +2Rm \right)}{m} \left(W_1(\mathcal{P}_R\# Q,\mathcal{P}_R\# P)+\varepsilon_m \right) + D_R^2\frac{\log\left( \frac{m}{\delta} \right)}{m} },
\end{multline*}
where $\varepsilon_m = \mathcal{O}\left(1\right)$ defined in \Cref{th: compact_mcall}.
As in \Cref{th: main_gaussian_lpz}, we have:
\[W_1(\mathcal{P}_R\# Q,\mathcal{P}_R\# P) \leq  W_1(Q,P)+2(M+1)\frac{\beta\sqrt{2\beta}}{m}. \]
We have
\begin{align*}
|\Delta_S(Q)| &\leq |\Delta_S(\mathcal{P}_R\# Q)| + |\Delta_S(Q) - \Delta_S(\mathcal{P}_R\# Q) |,
\intertext{And we have}
|\Delta_S(Q) - \Delta_S(\mathcal{P}_R\# Q) | & \leq \mathbb{E}_{h\sim Q}\left[|\Delta_S(h) - \Delta_S(\mathcal{P}_R(h))|\right] \\
& = \mathbb{E}_{h\sim Q}\left[|\Delta_S(h) - \Delta_S(\mathcal{P}_R(h))| \mathds{1}(||h||>R)\right] .
\intertext{And because $\ell$ is $K$-Lipschitz, $\Delta_S$ is $2K$-Lipschitz and we have:}
|\Delta_S(Q) - \Delta_S(\mathcal{P}_R\# Q) |&\leq 2K\mathbb{E}[||h-\mathcal{P}_R(h)||\mathds{1}(||h||>R)] \\
& \leq 2K\mathbb{E}[||h||\mathds{1}(||h||>R)].
\intertext{Finally, applying \Cref{l: gaussian_tail} gives:}
|\Delta_S(Q) - \Delta_S(\mathcal{P}_R\# Q) |&\leq 2K(M+1)\frac{\beta\sqrt{2\beta}}{m} = \mathcal{O}\left( \frac{1}{m} \right).
\end{align*}
Then we have:
\begin{multline}
\label{eq: complete_lpz_unbounded}
|\Delta_S(Q)| \leq 2K(M+1)\frac{\beta\sqrt{2\beta}}{m} + \\
\sqrt{ 2 K(2 K+D_R) \frac{\log(\frac{1}{\delta}) + 2d\log\left(1 +2Rm \right)}{m} \left(W_1( Q, P)+\alpha_m \right) + D_R^2\frac{\log\left( \frac{3m}{\delta} \right)}{m} },
\end{multline}
where $\alpha_m= \mathcal{O}\left(1 + \sqrt{\nicefrac{d\log(Rm)}{m}}\right)$ defined in \Cref{th: main_gaussian_lpz}.
Finally we exploit that $R= \mathcal{O}(\sqrt{d\log(d)},\sqrt{\log(m)})$ (cf. \Cref{rem: rad_rate}) and $D_R=\mathcal{O}(1+K^2R)$, to conclude the proof for all the three regimes.
\end{proof}

\subsection{PAC-Bayesian bounds for convex smooth functions}
\label{sec: main_sec_gaussian_smooth}

This section is focused on convex smooth loss functions, which are well suited for many optimisation objectives.
We show that under \textbf{(A2)}, it is possible to transform \Cref{th: compact_mcall} into a bound for smooth functions on all $\mathbb{R}^d$ when the loss remain bounded. We also show that it is possible to obtain a PAC-Bayesian bound for smooth unbounded loss functions.

\begin{theorem}
\label{th: main_gaussian_smooth}
Assume that $d\geq 3$, $\mathcal{H}= \mathbb{R}^d$ and that the loss satisfies \textbf{(A2)} and lies in $[0,1]$ over $\mathcal{H}$. For any $\delta>0, 0\leq \alpha\leq \beta, M\geq 0$, let $P\in C_{\alpha,\beta,M}$ a (data-free) prior distribution. Then, with probability $1-\delta$, for any posterior distribution $Q\in C_{\alpha,\beta,M}$:
\begin{align*}
|\Delta_S(Q)| \leq 2 \frac{\beta\sqrt{2\beta}}{m} + \sqrt{2D_R(2D_R+1)\frac{2d\log\left(3\frac{1 +2Rm }{\delta}\right)}{m} \left(W_1(Q,P)+ \alpha_m \right) + \frac{\log\left( \frac{3m}{\delta} \right)}{m} },
\end{align*}
with $R= \mathcal{O}\left( \max \sqrt{d\log(d)}, \sqrt{\log(m)}   \right)$,  $D_R= D+LR$ and $\alpha_m= \mathcal{O}(1)$ is defined in \Cref{th: main_gaussian_lpz}.
\end{theorem}
The key idea of the proof is to state that on a compact space, a smooth function is also Lipschitz. Therefore, the proof follows the same route as the one of \Cref{th: main_gaussian_lpz}, with additional technical steps. We then defer it to \Cref{sec: proof_smooth}.
We note that, even for bounded losses, the price to pay to consider smooth functions instead of Lipschitz ones is an extra factor $D_R= \mathcal{O}(1+R)$ when $D>0$. Therefore, in the general case we lose the idea that a tight smooth function will change the convergence rate of the problem as in general the upper bound $D$ of  $\sup_{z}|\ell(0_{\mathbb{R}^d},z)|$ is greater than zero. However, we are able to obtain results still useful when enough data is available. We also show it is possible to obtain a WPB bound for unbounded convex smooth functions.

\begin{corollary}
\label[corollary]{cor: unbounded_smooth}
Assume that $d\geq 3$, $\mathcal{H}= \mathbb{R}^d$ and that the (unbounded) loss satisfies \textbf{(A2)}. For any $\delta>0, 0\leq \alpha\leq \beta, M\geq 0$, we assume that $R>0$ is the smallest value satisfying \texttt{Rad}.
We assume that $\sup_{z\in\mathcal{Z}} || \ell(0,z)|| =D_\ell < +\infty$.
Let $P\in C_{\alpha,\beta,M}$ a (data-free) prior distribution.
Then, with probability $1-\delta$ , for any posterior distribution $Q\in C_{\alpha,\beta,M}$, the three following bounds holds.

\noindent \textbf{Low-data regime} $(d\geq m)$
\begin{align*}
|\Delta_S(Q)|  \leq\Tilde{\mathcal{O}}\left( \sqrt{\frac{d^{\nicefrac{5}{2}}}{m}\left(\sqrt{\frac{d}{m}}+ W_1(Q,P)\right)}   \right).
\end{align*}

\noindent \textbf{Transitory regime} $(d<m, d\log(d)\geq \log(m))$

\begin{align*}
|\Delta_S(Q)|  \leq\Tilde{\mathcal{O}}\left( \sqrt{\frac{d^{\nicefrac{5}{2}}}{m}\left(1+ W_1(Q,P)\right)}   \right).
\end{align*}

\noindent \textbf{Asymptotic regime} $(d\log(d)< \log(m))$
\begin{align*}
|\Delta_S(Q)|  \leq\Tilde{\mathcal{O}}\left( \sqrt{\frac{d}{m}}\left(1 + W_1(Q,P)\right)  \right).
\end{align*}
In all these bounds, $\Tilde{\mathcal{O}}$ hides a polynomail factor in $(\log(d),\log(m))$.
For a complete formulation of the bounds, we refer to \eqref{eq: complete_smooth_unbounded}.
\end{corollary}
We remark that this theorem is particularly interesting in the transitory and asymptotic regime as, contrary to \Cref{cor: unbounded_lpz}, we do not have a Lipschitz constant to attenuate the impact of the dimension (indeed we have $D_R = D + LR$ and in general $D>0$). However, this bound remains of great interest when many data are available as the smoothness assumption is often used in optimisation.

\begin{proof}[Proof of \Cref{cor: unbounded_smooth}]

Firstly, we use \Cref{th: compact_mcall} which state that for any prior on a compact, loss function $\ell\in  [0,1]$ being uniformly $K$-Lipschitz on this compact gives with probability at least $1-\delta$:
\begin{equation*}
|\Delta_S(Q)| \leq \sqrt{ 2K(2K+1) \frac{\log(\frac{3}{\delta}) + 2d\log\left(1 +2Rm \right)}{m} \left(W_1(Q,P)+\varepsilon_m \right) + \frac{\log\left( \frac{3m}{\delta} \right)}{m} }.
\end{equation*}
Let $P\in C_{\alpha,\beta,M}$. We fix $R$ to be the smallest value satisfying \texttt{Rad} and we assume \textbf{(A2)}.
On $\mathcal{B}(0,R)$, as seen in the proof of \Cref{th: main_gaussian_smooth}, $\ell$ is uniformly $D_R:= D+LR$-Lipschitz, so $\ell$ is bounded on this ball by $C_R:=D_\ell+RD_R= \mathcal{O}(1+R^2)$.
We apply \Cref{th: compact_mcall} on the loss function $\ell'= \ell/C_R$ and we multiply the resulting bound by $C_R$. Recall that $\ell'$ takes value in $[0,1]$  and is $D_R/C_R$-Lipschitz. We then have with high probability, for any $Q\in C_{\alpha,\beta,M}$:
\begin{multline*}
|\Delta_S(\mathcal{P}_R\# Q)| \\ \leq \sqrt{ 2D_R(2D_R+C_R) \frac{\log(\frac{3}{\delta}) + 2d\log\left(1 +2Rm \right)}{m} \left(W_1(\mathcal{P}_R\# Q,\mathcal{P}_R\# P)+\varepsilon_m \right) + C_R^2\frac{\log\left( \frac{3m}{\delta} \right)}{m} },
\end{multline*}
where $\varepsilon_m = \mathcal{O}\left(1\right)$ defined in \Cref{th: compact_mcall}.
As in \Cref{th: main_gaussian_lpz}, we have:
\[W_1(\mathcal{P}_R\# Q,\mathcal{P}_R\# P) \leq  W_1(Q,P)+2(M+1)\frac{\beta\sqrt{2\beta}}{m}. \]
We have:
\begin{align*}
|\Delta_S(Q)| &\leq |\Delta_S(\mathcal{P}_R\# Q)| + |\Delta_S(Q) - \Delta_S(\mathcal{P}_R\# Q) |,
\intertext{And we have:}
|\Delta_S(Q) - \Delta_S(\mathcal{P}_R\# Q) | & \leq \mathbb{E}_{h\sim Q}\left[|\Delta_S(h) - \Delta_S(\mathcal{P}_R(h))|\right] \\
& = \mathbb{E}_{h\sim Q}\left[|\Delta_S(h) - \Delta_S(\mathcal{P}_R(h))| \mathds{1}(||h||>R)\right] .
\intertext{We study the last gap more carefully:}
|\Delta_S(h) - \Delta_S(\mathcal{P}_R(h))|& = \mathbb{E}_z[|\ell(h,z)- \ell(\mathcal{P}_R(h),z)|] + \frac{1}{m}\sum_{i=1}^m |\ell(h,z_i)- \ell(\mathcal{P}_R(h),z_i)|.
\intertext{And we know that for any $z$, because $\ell$ is convex smooth:}
\ell(h,z)- \ell(\mathcal{P}_R(h),z) &\leq \nabla_h \ell(\mathcal{P}_R(h),z)^T(h-\mathcal{P}_R(h)) + \frac{L}{2}||h-\mathcal{P}_R(h)||^2 || \\
& \leq D_R||h- \mathcal{P}_R(h)|| + \frac{L}{2}||h-\mathcal{P}_R(h)||^2 ||.
\intertext{We also have by convexity:}
\ell(\mathcal{P}_R(h),z) - \ell(h,z) &\leq \nabla_h\ell(\mathcal{P}_R(h),z)^T(\mathcal{P}_R(h)-h) \\
& \leq D_R ||h-\mathcal{P}_R(h)||.
\intertext{In any case, we have for any $h,z$: }
|\ell(h,z)- \ell(\mathcal{P}_R(h),z)| &\leq D_R||h-\mathcal{P}_R(h)|| + \frac{L}{2}||h-\mathcal{P}_R(h)||^2.
\intertext{Thus:}
|\Delta_S(Q) - \Delta_S(\mathcal{P}_R\# Q) | & \leq D_R\mathbb{E}_{h\sim Q}\left[||h-\mathcal{P}_R(h)||\mathds{1}(||h||>R)\right] \\
& + \frac{L}{2}\mathbb{E}_{h\sim Q}\left[||h-\mathcal{P}_R(h)||^2\mathds{1}(||h||>R)\right] \\
& \leq D_R\mathbb{E}_{h\sim Q}\left[||h||\mathds{1}(||h||>R)\right] + \frac{L}{2}\mathbb{E}_{h\sim Q}\left[||h||^2\mathds{1}(||h||>R)\right].
\intertext{And thanks to \Cref{l: gaussian_tail}, we finally have:}
|\Delta_S(Q) - \Delta_S(\mathcal{P}_R\# Q) | & \leq \left( D_R + \frac{L}{2}(M+1) \right)(M+1)\frac{\beta\sqrt{\beta}}{m}.
\end{align*}
Then we have:
\begin{multline}
\label{eq: complete_smooth_unbounded}
|\Delta_S(Q)| \leq \left( D_R + \frac{L}{2}(M+1) \right)(M+1)\frac{\beta\sqrt{\beta}}{m} + \\
\sqrt{ 2D_R(2D_R+C_R) \frac{\log(\frac{3}{\delta}) + 2d\log\left(1 +2Rm \right)}{m} \left(W_1(Q, P)+\alpha_m \right) + C_R^2\frac{\log\left( \frac{3m}{\delta} \right)}{m} },
\end{multline}
where $\alpha_m= \mathcal{O}\left(1 + \sqrt{\nicefrac{d\log(Rm)}{m}}\right)$ defined in \Cref{th: main_gaussian_lpz}.
Finally, we exploit that $R= \mathcal{O}(\sqrt{d\log(d)},\sqrt{\log(m)})$ (cf \Cref{rem: rad_rate}), that $D_R=\mathcal{O}(1+R)$ and $C_R=\mathcal{O}(1+R^2)$, to conclude the proof for all the three regimes.
\end{proof}

\section{Wasserstein PAC-Bayes with data-dependent priors}
\label{sec: data_dep_priors}

In PAC-Bayes learning, obtaining results holding with data-dependent priors is a widely studied topic. The reason behind that is that it is more meaningful to compare the posterior distribution, usually obtained via an optimisation procedure to a competitive one (classically the Gibbs posterior) to ensure tight generalisation bounds.
A classical way to do so is to use differential privacy as in \citet{dziugaite2018data}. However, their contribution relies on bounded losses to apply the \emph{exponential mechanism}, a useful tool to determine whether an algorithm is differentially private. We exploit new theorems from \citet{minami2016diff,rogers2016max} which allow us to exploit differentially private priors when the loss is unbounded, convex and Lipschitz. We recall in \Cref{sec: back_dp} elements of differential privacy.

\textit{A PAC-Bayesian bound for Lipschitz convex losses with data-dependent prior.} We now state a PAC-Bayes theorem valid for differentially private probability kernels. The proof elaborates on \citet[Theorem 4.2]{dziugaite2018data} and is based on the following bound, which is a minor modification of \eqref{eq: complete_lpz_unbounded}, making it valid for any prior (and not only Gaussian ones).

\begin{theorem}
\label{th: data_dep}
Assume that $d\geq 3$, $\mathcal{H}= \mathbb{R}^d$ and that the loss is convex and satisfies \textbf{(A1)}. Let $\beta_m= \mathcal{O}(\nicefrac{1}{\sqrt{m}})$ and $\lambda \leq \sqrt{m}$.
Let $P\in C_{\alpha,\beta,M}$ a (data-free) prior distribution. Then, for any $\beta_m<\delta<1$, with probability $1-\delta$ , for any posterior distribution $Q\in C_{\alpha,\beta,M}$ and the Gibbs prior $P_{-\frac{\lambda}{2K} R_S}$, the following bound holds.

\noindent\textbf{Low-data regime} $(d\geq m)$
\begin{align*}
|\Delta_S(Q)|  \leq\Tilde{\mathcal{O}}\left( \sqrt{2K\frac{d^{\nicefrac{3}{2}}}{m}\left(\sqrt{\frac{d}{m}}+  W_1(Q,P_{-\frac{\lambda}{2K}R_S}) +f_R\left(P_{-\frac{\lambda}{2K}R_S}\right)\right)+(1+K^2d)\frac{\log\left( \frac{m}{\delta} \right)}{m}}   \right).
\end{align*}
\textbf{Transitory regime} $(m>d,\; d\log(d)\geq \log(m))$

\begin{align*}
|\Delta_S(Q)|  \leq\Tilde{\mathcal{O}}\left( \sqrt{2K\frac{d^{\nicefrac{3}{2}}}{m}\left(1+ W_1(Q,P_{-\frac{\lambda}{2K}R_S})+ f_R\left(P_{-\frac{\lambda}{2K}R_S}\right)\right)+(1+K^2d)\frac{\log\left( \frac{m}{\delta} \right)}{m}}   \right).
\end{align*}
\textbf{Asymptotic regime} $(d\log(d)< \log(m))$
\begin{align*}
|\Delta_S(Q)|  \leq\Tilde{\mathcal{O}}\left( \sqrt{2K\frac{d}{m}\left(1 + W_1(Q,P_{-\frac{\lambda}{2K}R_S})\right)+(1+K^2\log(m))\frac{\log\left( \frac{m}{\delta} \right)}{m}}   \right),
\end{align*}
where $R= \mathcal{O}\left( \max \sqrt{d\log(d)}, \sqrt{\log(m)}   \right)$, $f_R(P) := W_1(\mathcal{P}_R\#P,P)$.
In the above $\Tilde{\mathcal{O}}$ hides a polynomial dependency in $(\log(d),\log(m))$. For an explicit formulation of the bounds, we refer to \eqref{eq: complete_data_dep}.
\end{theorem}
Note that in the asymptotic bound, the condition to get rid of $f_R(P_{-\frac{\lambda}{2K}R_S})$ is that $\lambda$ is a fixed constant, in particular it does not depend on $m$. This is essential to apply the law of large numbers: a fixed learning rate in the Gibbs posterior is required for a bound with only explicit terms.
Furthermore, an important message is that Lipschitz functions are well suited to the PAC-Bayes framework through Wasserstein distances. Indeed, not only are we able to recover McAllester or Catoni-type WPB bounds, but we also obtain WPB with data-dependent priors using the same techniques than PAC-Bayes learning with KL divergences. Data-dependent WPB bounds have also an additional benefit as they provide guarantees for the Bures-Wasserstein SGD of \citet{lambert2022variational} as shown in \Cref{sec: gene_sgd}.

\begin{proof}[Proof of \Cref{th: data_dep}]
Firstly, we start from a slightly modified version of \Cref{eq: complete_lpz_unbounded} which holds for any prior distribution (and not only Gaussian ones).
To obtain it we restart from the triangle inequality $W_1(\mathcal{P}_R\#Q, \mathcal{P}_R\#P ) \leq W_1(\mathcal{P}_R\#Q,Q) + W_1(Q,P) + f_R(P)$. where $f_R(P) := W_1(\mathcal{P}_R\#P,P)$ and we apply exactly the same route of proof than in \Cref{cor: unbounded_lpz}. We then obtain, for any data-free prior $P$, with probability at least $1-\delta$, for any $Q\in C_{\alpha,\beta,M}$:
\begin{multline*}
|\Delta_S(Q)| \leq 2K(M+1)\frac{\beta\sqrt{2\beta}}{m} + \\
\sqrt{ C_R \frac{\log(\frac{1}{\delta}) + 2d\log\left(1 +2Rm \right)}{m} \left(W_1( Q, P)+\alpha_m  + f_R(P)\right) + D_R^2\frac{\log\left( \frac{m}{\delta} \right)}{m} },
\end{multline*}
where $D_R= D+KR$ and $C_R= 2K(2K+D_R)$ ($D,K$ defined in \textbf{(A1)}).
We then denote by $\texttt{Bound}(S,P,Q,\delta)$ the bound:
\begin{multline*}
|\Delta_S(Q)| > 2K(M+1)\frac{\beta\sqrt{2\beta}}{m} + \\
\sqrt{ C_R \frac{\log(\frac{1}{\delta}) + 2d\log\left(1 +2Rm \right)}{m} \left(W_1( Q, P)+\alpha_m  + f_R(P)\right) + D_R^2\frac{\log\left( \frac{m}{\delta} \right)}{m} }.
\end{multline*}
And for a given $\delta'$, let $\texttt{Ev}(P,\delta'):= \{S \in \mathcal{Z}^m \mid  \exists Q \in C_{\alpha,\beta,M} \text{ s.t. } \texttt{Bound}(S,P,Q,\delta') \text{ holds} \}$.
We know that for a data-free prior $P$, $\mathbb{P}_{S\in\mu^m}(S\in \texttt{Ev}(P)) \leq \delta$.
To exploit the differential privacy framework, we first assume having a differentially private probability kernel $\mathcal{P}$. We fix $\beta>0$ and re-exploit the idea of \citet{dziugaite2018data}:
\begin{align}
\label{eq: pac_b_diff_priv}
\mathbb{P}_{S \sim \mu^m}\{S \in \texttt{Ev}(\mathcal{P}(S),\delta')\} & \leq \mathrm{e}^{\mathbf{I}_{\infty}^\beta(\mathcal{P} ; m)} \underset{\left(S, S^{\prime}\right) \sim \mu^{2m} }{\mathbb{P}}\left\{S \in \texttt{Ev}\left(\mathcal{P}\left(S^{\prime}\right)\right)\right\}+\beta \\
& \leq \mathrm{e}^{\mathbf{I}_{\infty}^\beta(\mathcal{P} ; m)} \delta^{\prime}+\beta = \delta .
\end{align}
The last line holds for any $\delta > \beta$ by fixing $\delta^{\prime}=\mathrm{e}^{-I_{\infty}^\beta(\mathcal{P} ; m)}(\delta-\beta)$.
Note that $\log\left(\nicefrac{1}{\delta'}   \right) = \log\left(\nicefrac{1}{\delta - \beta}   \right) + I_{\infty}^\beta(\mathcal{P} ; m)$, this suggests to bound the $\beta$-approximate max-information. To do so, we need to give specific values for the pair $(\varepsilon,\gamma)$.
More concretely, let $\varepsilon= \sqrt{\nicefrac{\log(m)}{m}}, \gamma= \nicefrac{\varepsilon}{m^4}$.
Then thanks to \Cref{prop: rogers}, we know that for $\beta_m := \mathcal{O}(\nicefrac{1}{m})$, we have:
\begin{equation}
\label{eq: rogers_bound}
I_{\infty}^\beta(\mathcal{P}, m)=O\left(\log(m) \right) .
\end{equation}
The last thing to do is to prove that the probability kernel $\mathcal{P}_0(S):= P_{-\lambda' m R_S}$ is $(\varepsilon,\gamma)$ differentially private. This is true thanks to \Cref{prop: minami} which states that $\mathcal{P}_0$ satisfies differential privacy as long as $\lambda' \leq \lambda_{m}$ with:
\begin{equation}
\label{eq: inv_temp}
\lambda_m:= \frac{1}{2K}\sqrt{\frac{\alpha \log(m)}{m\left(1-2\log\log(m) +10\log(m) \right)}} = \mathcal{O}\left(\frac{1}{\sqrt{m}}\right).
\end{equation}
Note that $\alpha$ intervenes because for any prior $P\in C_{\alpha,\beta,M}$, $-\log P(.)$ is $\alpha$-strongly convex.
From now we consider $\lambda'= \frac{\lambda}{2Km}$ where $\lambda \leq \sqrt{m}$. We then have $\lambda' \leq \lambda_m$.
We then know, thanks to \Cref{eq: pac_b_diff_priv} with $\beta=\beta_m$, that for any $\delta > \beta_m$,  $\mathbb{P}_{S \sim \mu^m}\{S \in \texttt{Ev}(\mathcal{P}_0(S),\delta') \leq \delta$
with $\delta^{\prime}=\mathrm{e}^{-I_{\infty}^\beta(\mathcal{P} ; m)}(\delta-\beta)$
Taking the complementary event and recalling that thanks to \Cref{eq: rogers_bound}, $\log\left(\nicefrac{1}{\delta'}   \right) = \log\left(\nicefrac{1}{\delta - \beta_m}   \right) + \mathcal{O}(\log(m))$ gives,
for any data-free Gaussian prior $P$, for any $\delta > \beta_m$, with probability at least $1-\delta$, for any $Q\in C_{\alpha,\beta,M}$:
\begin{multline}
\label{eq: complete_data_dep}
|\Delta_S(Q)| \leq 2K(M+1)\frac{\beta\sqrt{2\beta}}{m} \\
+\sqrt{ C_R \frac{\log(\frac{1}{\delta-\beta_m}) + \mathcal{O}(\log(m)) +2d\log\left(1 +2Rm \right)}{m} \left(W_1( Q, P_{-\frac{\lambda}{2K} R_S})+\alpha'_m + f_R(P_{-\frac{\lambda}{2K} R_S}) \right) }  \\
+\sqrt{ D_R^2\frac{\log\left( \frac{m}{\delta-\beta_m} \right) +\mathcal{O}(\log(m)) }{m}},
\end{multline}
where $\alpha'_m= \mathcal{O}(1+ \nicefrac{d\log(m)}{m})$ has the same analytical expression than $\alpha_m$ (defined in \Cref{th: main_gaussian_lpz}) but where all the occurences of $\delta$ have been replaced by $\delta'$.
Note that in the last equation, we used $\sqrt{a+b} \leq \sqrt{a} + \sqrt{b}$ ($a,b>0$) for the sake of readability but we put everything within the same square root in our theorem as it is tighter.
Then, exploiting that $R= \mathcal{O}(\sqrt{d\log (d)}, \sqrt{\log(m)})$, gives us the results for the low-data and transitory regimes.
\medskip

\noindent Also, we are able to prove that asymptotically, because $R \sqrt{\log(m)}\rightarrow \infty$ when $m$ goes to infinity:
$$f_R(P_{-\frac{\lambda}{2K} R_S}) \leq \mathbb{E}[||X-\mathcal{P}_R(X)||] \underset{m\rightarrow \infty}{\rightarrow} 0,    $$
where $X$ follows the Gibbs distribution $P_{-\frac{\lambda}{2K} R_S}$. The convergence to zero comes from the dominated convergence theorem.
Indeed,
\[ \mathbb{E}[||X-\mathcal{P}_R(X)||] = \int_{\mathbb{R}^d} g_m(x) \mathrm{d}P(x), \]
with $g_m(x)= ||x - \mathcal{P}_R(x)||\frac{\exp\left(-\lambda R_S(x) \right)}{\mathbb{E}_P[\exp\left(-\lambda R_S(x) \right)]}$. Thus, bounding crudely gives:
\[  \mathbb{E}[||X-\mathcal{P}_R(X)||] \leq \frac{1}{\inf_{m\geq 1} \mathbb{E}_P[\exp\left(-\lambda R_S(x) \right)]} \int_{\mathbb{R}^d} ||x - \mathcal{P}_R(x)|| \mathrm{d}P(x).\]
We know that $\inf_{m\geq 1} \mathbb{E}_P[\exp\left(-\lambda R_S(x) \right)] := \inf_{m\geq 1} \mathbb{E}_P[f_m(x) ] >0$ because $f_m$ is $\lambda K$ - lipschitz ($x \rightarrow e^{- \lambda x}$ is $\lambda$-lipschitz and the loss $\ell$ is $K$-lipschitz)
and converges almost surely on $\mathbb{R^d}$ towards $x\rightarrow \exp{-\lambda R(x)}$. Indeed, thanks to the law of large numbers, we know that on $\mathbb{Q}^d$, $f_m \rightarrow f$ almost surely and using that all the sequence is $\lambda K$ lipschitz extends the result to all $\mathbb{R}^d$.
We also notice that for any $m$, $f_m \leq 1$ so we can use the dominated convergence theorem to conclude that $\mathbb{E}_P[f_m(x)] \rightarrow  \mathbb{E}_P[\exp(-\lambda R(X)) ] >0.$ So $\inf_{m\geq 1} \mathbb{E}_P[\exp\left(-\lambda R_S(x) \right)]>0$.
The last thing to do is to use \Cref{l: gaussian_tail} to ensure that $\int_{\mathbb{R}^d} ||x - \mathcal{P}_R(x)|| \mathrm{d}P(x) \rightarrow 0$.
This allows us to get rid of $f_R$ for the asymptotic regime and then, conclude the proof.
\end{proof}

\section{Generalisation ability of the Bures-Wasserstein SGD}

\label{sec: gene_sgd}

For the sake of completeness, we recall (and precise) several elements already defined in \Cref{sec: intro_optim}.
In PAC-Bayes learning, the following learning algorithm can be derived from a relaxation of \citet[][Theorem 1.2.6]{catoni2007pac}, for any data-free prior $P$ and inverse PAC-Bayesian temperature $\lambda>0$:

\[ \underset{Q\in \mathcal{M}_1(\mathcal{H})}{\operatorname{argmin}} \mathbb{E}_{h\sim Q}[R_S(h)] + \frac{\operatorname{KL}(Q,P)}{\nicefrac{\lambda}{2K}}.  \]

\noindent We considered the parameter $\nicefrac{\lambda}{2K}$ as it was suggested by \Cref{th: data_dep}.
A closed form solution is given by the Gibbs posterior $Q^*:= P_{-\nicefrac{\lambda}{2K}}$ such that $\mathrm{d}Q^* \propto \exp(-V_S(h))\mathrm{d}h$, with $V_S(h) = \frac{\lambda}{2K}R_S(h) - \log(\mathrm{d}P(h))$ and $\mathrm{d}h$ being the Lebesgue measure.
However, such a measure can be difficult to estimate in practice. Two solutions are available. We can estimate the Gibbs posterior through MCMC methods that rely on Markov chains which (approximately) converge to $Q^*$. However, there is no clear stopping criterion to obtain a good approximate of the true posterior. Otherwise, we can exploit variational inference (VI) to produce rapidly a basic yet informative summary statistics on a subclass of $\mathcal{M}_1(\mathcal{H})$.
In this section, we focus on the VI approach. As $Q^*$ is the result of an optimal trade-off between the empirical loss $R_S$ and the $KL$ divergence (weighed by $\lambda$) acting as a regulariser, we consider the closest measure of $\operatorname{BW}(\mathbb{R}^d)$ from $Q^*$ with respect to the KL divergence:
\[ \hat{Q} = \mathcal{N}(\hat{m},\hat{\Sigma}) := \underset{Q\in \operatorname{BW}(\mathbb{R}^d)}{\operatorname{argmin}} \operatorname{KL}(Q,Q^*). \]
At the cost of this approximation, can we have an optimisation algorithm with convergence guarantees which goes to $\hat{Q}$? Furthermore, if enough data is available, does $\hat{Q}$ possess a good generalisation ability?
We first state the assumptions holding throughout the whole section.

\textit{(A3):} We assume that $\mathcal{H}=\mathbb{R}^d$ and
\begin{itemize}
  \item There exists $M>0$ such that $||\hat{m}|| \leq M$ almost surely.
  \item $\ell$ is twice differentiable, and \textbf{(A1), (A2)} hold. In particular, $\ell$ is $L$-smooth, convex and uniformly $K$-Lipschitz over $\mathcal{H}$. We furthermore assume that $L=1$.
  \item The prior $P$ used in the definition of $Q^*$ is a Gaussian with mean $0$ and covariance matrix $\Sigma= \text{diag}(\gamma), 1\geq\gamma>0$. We assume $\lambda \leq 2K$ in the definition of $Q^*$.
\end{itemize}
Note that under \textbf{(A3)}, we have $0 \prec \alpha I \preceq \nabla^2 V_S \preceq I$.
The work of \citet[Theorem 4]{lambert2022variational} provides convergence guarantees for SGD over the Bures-Wasserstein space when \textbf{(A3)} holds (in particular, they do not even requires the uniformly Lipschitz assumption). We first state their algorithm in \Cref{alg: sgd}.
\begin{algorithm}[ht]
\SetAlgoLined
\SetKwInOut{Initialisation}{Initialisation}
\SetKwInOut{Parameter}{Parameters}
\Parameter{Strong convexity parameter $\alpha>0$, radius $M>0$; step size $\eta>0$, initial mean $m_0$, initial covariance $\Sigma_0$}
Set up $\hat{Q}_0 = \mathcal{N}(m_0,\Sigma_0)$. \\
\For{$k= 0..N-1$}{
Draw a sample $X_k\sim \hat{Q}_k$.\\
Set $m_k^+ = m_k - \eta \nabla V_S(X_k)$.\\
Set $M_k= I- \eta (\nabla V^2(X_k) -\Sigma_k^{-1}) $.\\
Set $\Sigma_k^+ = M_k\Sigma_k M_k$.\\
Set $m_{k+1} = \mathcal{P}_{M}(m_k^+),\;\Sigma_{k+1} = \operatorname{clip}^{1/\alpha}\Sigma_k^+$.\\
Set $\hat{Q}_{k+1}= \mathcal{N}(m_{k+1},\Sigma_{k+1})$
}
\textbf{Return} $(\hat{Q}_k)_{k=1...N}$.
\caption{Bures-Wasserstein SGD.}
\label{alg: sgd}
\end{algorithm}
Note that \Cref{alg: sgd} is a slight adaptation of the work of \citet{lambert2022variational}. Indeed, we added a projection step $\mathcal{P}_M$  within the compact of radius $M$ in $\mathbb{R}^d$. This does not change the convergence guarantees stated in \Cref{th: lambert} as long as we assume \textbf{(A3)}.

\begin{theorem}
\label{th: lambert}
Assume \textbf{(A3)}. Also, assume that $\eta \leq \frac{\alpha^2}{60}$ and that we initialize \Cref{alg: sgd} at a matrix satisfying $\frac{\alpha}{9} I \preceq \Sigma_{0} \preceq \frac{1}{\alpha} I$. Then, for all $k \in \mathbb{N}$,
$$
\mathbb{E} W_2^2\left(\hat{Q}_k, \hat{Q}\right) \leq \exp (-\alpha k \eta) W_2^2\left(\hat{Q}_0, \hat{Q}\right)+\frac{36 d \eta}{\alpha^2} .
$$
In particular, we obtain $\mathbb{E} W_2^2\left(\hat{Q}_k, \hat{Q}\right) \leq \varepsilon^2$ provided we set $h \asymp \frac{\alpha^2 \varepsilon^2}{d}$ and the number of iterations to be $k \gtrsim \frac{d}{\alpha^3 \varepsilon^2} \log \left(W_2\left(\hat{Q}_0, \hat{Q}\right) / \varepsilon\right)$.
\end{theorem}
We want to incorporate \Cref{th: lambert} within \Cref{th: data_dep}. To do so, we need to make sure that the outputs of \Cref{alg: sgd} and $\hat{Q}$ lie a compact of $BW(\mathbb{R}^d)$. To do so we exploit the following lemma, which sums up the work of \citet{lambert2022variational} (namely their Lemma 6 and the discussion in Section 3.3). 

\begin{lemma}
\label[lemma]{l: measures_in_compact}
Assume \textbf{(A3)} and the step-size $\eta$ of \Cref{alg: sgd} is lesser than $\frac{\alpha^2}{60}$. Also in \Cref{alg: sgd}, assume that $\frac{\alpha}{9} I \preceq \Sigma_k$.
Then $\frac{\alpha}{9} I \preceq\Sigma_{k}^+$, and so, $\frac{\alpha}{9}I \preceq\Sigma_{k+1} \preceq \frac{1}{\alpha} I$.
Furthermore, $I \preceq \hat{\Sigma}  \preceq \frac{1}{\alpha} I$.
Thus, if the initialisation of \Cref{alg: sgd} is such that $\frac{\alpha}{9} I \preceq \Sigma_{0} \preceq \frac{1}{\alpha} I$, then the sequence $(\hat{Q}_k)_{k\geq 0}$ and $\hat{Q}$ are in the compact $C_{\frac{\alpha}{9}, \frac{1}{\alpha}, M}$.
\end{lemma}
Using \Cref{l: measures_in_compact}, we now can apply \Cref{th: data_dep} and obtain the main result of this section.
\begin{theorem}
\label{th: main_sgd}
Assume \textbf{(A3)}, also assume that $d\geq 3$. Let $\beta_m= \mathcal{O}(\nicefrac{1}{\sqrt{m}})$ and fix any $\beta_m<\delta<1$.
Assume that we perform \Cref{alg: sgd}, with step size $\eta \asymp \frac{\alpha^2 \delta}{d}$ and the number of iterations to be $N \gtrsim \frac{d}{\alpha^3 \delta} \log \left(W_2\left(Q_0, \hat{Q}\right) / \delta\right)$.
We also set the initialisation such that $\frac{\alpha}{9} I \preceq \Sigma_{0} \preceq \frac{1}{\alpha} I$,
then we can upper bound the generalisation ability of $\hat{Q}_N$, with probability $1-2\delta$:

\noindent \textbf{Asymptotic regime} $(d\log(d)< \log(m))$
\begin{align*}
|\Delta_S(\hat{Q}_N)|  \leq\Tilde{\mathcal{O}}\left( \sqrt{2K\frac{d}{m}\left(1 + W_1(\hat{Q},Q^*)\right)+ (1+K^2\log(m)) \frac{\log\left( \frac{m}{\delta} \right)}{m}} \right),
\end{align*}
where $\Tilde{\mathcal{O}}$ hides a polynomial dependency in $(\log(d),\log(m))$. We refer to \eqref{eq: complete_bound_sgd} for a bound presenting the explicit influence of the Bures-Wasserstein SGD.
\end{theorem}
\Cref{th: main_sgd} is based on \Cref{eq: complete_bound_sgd} which answers the question stated in the 'Our aims in this paper' paragraph of \Cref{sec: intro_optim}. We successfully designed a bound of the form of \eqref{eq: wanted_pattern} by incorporating the optimisation guarantees of \citet{lambert2022variational} onto a statistical framework.
As such, this bound is a bridge between optimisation and PAC-Bayes learning. To the best of our knowledge, it is the first time that PAC-Bayes is able to explain why the minimiser attained by an optimisation procedure on a measure space is also able to generalise well. Until now PAC-Bayes guarantees were used as a check-in procedure, which means that during the optimisation phase it is possible to see whether the candidate predictor is able to generalise well. On the contrary our bound higlights, before any training, that the output of the Bures-Wasserstein SGD will become better at generalising, with the limit rate of $\sqrt{\frac{Kd}{m}W_1(\hat{Q},Q^*) + \frac{\log(m)}{m}}$.
\medskip

Let us analyse the bound: the convergence rate depends on the quality of the approximation $\hat{Q}$ of $Q^*$, this says that if Gaussian measures are not suited to approximate well the Gibbs posterior, then we sacrifice some generalisation ability. However this term is also controlled by the Lipschitz constant $K$: if $K$ is small, then the learning problem is easy enough to compensate both the curse of dimensionality and a possibly bad approximation $\hat{Q}$ of $Q^*$.
Again, the limit convergence rate is the statistical ersatz $\mathcal{O}\left( \sqrt{\nicefrac{\log(m)}{m}} \right)$. This roughly says that we cannot hope to converge better than a Hoeffding test bound in this setting. Finally note also that the step $\eta$ of \Cref{alg: sgd} now depends on $\delta$: this suggests that the Bures-Wasserstein SGD needs to be tuned with a smaller step size to ensure not only convergence, but also a good generalisation ability.
\begin{proof}[Proof of \Cref{th: main_sgd}]
We start from \Cref{th: data_dep}, considering the asymptotic case. We have with probability $1-\delta$, for the posterior $\hat{Q}_N$ obtained after $N$ steps of \Cref{alg: sgd} distribution $Q\in C_{\alpha,\beta,M}$ and the prior $Q^*$:
\begin{align*}
|\Delta_S(\hat{Q}_N)|  \leq\Tilde{\mathcal{O}}\left( \sqrt{2K\frac{d}{m}\left(1 + W_1(\hat{Q}_N,Q^*)\right)+ (1+K^2\log(m))\frac{\log\left( \frac{m}{\delta} \right)}{m}}   \right).
\end{align*}
Then, the triangle inequality gives that $W_1(\hat{Q}_N,Q^*) \leq W_1(\hat{Q}_N,\hat{Q}) + W_1(\hat{Q},Q^*)$.
Finally, we exploit \Cref{th: lambert} as follows:
\begin{align*}
W_1(\hat{Q}_N,\hat{Q}) & \leq \sqrt{W_2^2(\hat{Q}_N,\hat{Q})} & \text{by Jensen} \\
& \leq \sqrt{2\frac{\mathbb{E}[W_2^2(\hat{Q}_N,\hat{Q})]}{\delta}} & \text{by Markov}\\
& \leq \sqrt{2\frac{\exp (-\alpha N \eta) W_2^2\left(\hat{Q}_0, \hat{Q}\right)+\frac{36 d \eta}{\alpha^2}}{\delta}} & \text{by \Cref{th: lambert}.}
\end{align*}
Note that in the last line, we were able to apply \Cref{th: lambert} thanks to \Cref{l: measures_in_compact}.
This leads to the following bound:
\begin{equation}
\label{eq: complete_bound_sgd}
|\Delta_S(\hat{Q}_N)|  \leq \Tilde{\mathcal{O}}\left( \sqrt{2K\frac{d}{m}\left( f(N,\eta)\sqrt{W_2^2(\hat{Q}_0,\hat{Q})} + 1 +
\varepsilon \right)+ (1+K^2\log(m))\frac{\log\left( \frac{m}{\delta} \right)}{m}}   \right),
\end{equation}
where $f(N,\eta)= \sqrt{\frac{\exp (-\alpha N \eta) W_2^2\left(\hat{Q}_0, \hat{Q}\right)}{\delta}}$ and $\varepsilon= \sqrt{\frac{36 d \eta}{\alpha^2\delta}} + W_1(\hat{Q},Q^*)$.
Finally, using that with step size $\eta \asymp \frac{\alpha^2 \delta}{d}$ and the number of iterations to be $N \gtrsim \frac{d}{\alpha^3 \delta} \log \left(W_2\left(\hat{Q}_0, \hat{Q}\right) / \delta\right)$ allows us to bound:
$\sqrt{2\frac{\exp (-\alpha k \eta) W_2^2\left(\hat{Q}_0, \hat{Q}\right)+\frac{36 d \eta}{\alpha^2}}{\delta}} \leq 1$.
This concludes the proof.
\end{proof}

\section{Conclusion}

We extended the Wasserstein PAC-Bayes theory beyond the results of \citet{amit2022ipm}. We exploited optimisation results to explain the generalisation ability of existing algorithms and we instantiated this for the Bures-Wasserstein algorithm of \citet{lambert2022variational}. We conclude by discussing avenues for future works.

\textit{Can we exploit WPB for neural networks?} As shown in \Cref{fig: overview}, we had to assume, Lipschitzness, smoothness and convexity to reach \Cref{th: main_sgd}. Those assumptions are necessary in the current framework and to obtain the results of \citet{lambert2022variational} and thus, do not cover the important case of neural networks.
Therefore, an interesting lead to investigate would be to first, avoid smoothness to reach convex neural networks \citet{bengio2005convex} and also avoid the convexity assumption to reach the broader subclass of Lipschitz neural networks (\emph{e.g} \citealp{gouk2021regularisation}).
The case of Lipschitz neural networks is particularly interesting as WPB theory shows that a small Lipschitz constant is enough to attenuate the impact of dimensionality. Whether Lipschitzness is enough to reach WPB bound with data-dependent prior (\Cref{th: data_dep} requires an additional convexity assumption) might involve further results from differential privacy and is left for future work.

\textit{Are the classical PAC-Bayesian techniques suited to WPB?} In \Cref{th: compact_catoni,th: compact_mcall}, we exploited a surrogate of the change of measure inequality to then exploit the PAC-Bayesian theory. However, those techniques are developed around the control of an exponential moment which appears naturally through the change of measure inequality. The surrogate is tighter as it directly involves the true moment with respect to the prior: an interesting direction would be to check whether tighter concentration bounds(or other bounds exploiting weaker assumptions than a bounded loss) are accessible. Furthermore, we exploited covering numbers to state that, with high probability, the loss is close to a Lipschitz one. Those covering numbers, while crucial, involve explicitly the dimension of the problem. This is challenging as such a dependency do not appear explicitly in KL-based PAC-Bayes learning (although they play a role in the KL term). Whether covering numbers are essential to WPB learning is an open question.

%% file: appendix.tex

\section{Additional background}
\label{sec: background}

\subsection{Background on optimal transport and covering numbers}
\label{sec: back_compact}

We recall a basic property on covering numbers.

\begin{proposition}
\label[prop]{prop: covering}
For any $R,\varepsilon$, $N(\bar{\mathcal{B}}(0,R)), \varepsilon) \leq \left(1+\frac{2R}{\varepsilon}\right)^d$.
\end{proposition}
The following theorem is initially stated in \citep[Theorem 5.10]{villani2009optimal}.
\begin{theorem}[Kantorovich duality]
\label{th: kanto_dual}
Let $(\mathcal{X}, Q)$ and $(\mathcal{Y}, P)$ be two Polish probability spaces and let $c: \mathcal{X} \times \mathcal{Y} \rightarrow \mathbb{R} \cup\{+\infty\}$ be a lower semicontinuous cost function, such that
$$
\forall(x, y) \in \mathcal{X} \times \mathcal{Y}, \quad c(x, y) \geq a(x)+b(y)
$$
for some real-valued upper semicontinuous functions $a \in L^1(Q)$ and $b \in L^1(P)$. Then there is duality:
$$
\begin{aligned}
\min _{\pi \in \Pi(Q, P)} & \int_{\mathcal{X} \times \mathcal{Y}} c(x, y) d \pi(x, y) & =\sup _{\substack{(\psi, \phi) \in L^1(Q) \times L^1(P)\\ \phi-\psi \leq c}}\left[\int_{\mathcal{Y}} \phi(y) dP(y)-\int_{\mathcal{X}} \psi(x) dQ(x)\right],
\end{aligned}
$$
where $L_1(P)$ refers to the set of all functions integrable with respect to $P$ and the condition $\phi-\psi \leq c$ means that for all $x,y \in \mathcal{X}\times \mathcal{Y}, \phi(y)-\psi(x) \leq c(x,y)$.
\end{theorem}

\subsection{Technical background for \Cref{sec: wpb_gauss}}
\label{sec: background_gaussian}

The theorems of \Cref{sec: wpb_gauss} all rely on a well-chosen radius $R$ (seen here as an hyperparameter) verifying the following set of (non-restrictive) assumptions.

\textit{The set of assumptions \texttt{Rad}.}

We say that $R>0$ is satisfying $\texttt{Rad}(\alpha,\beta,M,m,d)$ (abbreviated as \texttt{Rad} when clear from context) for $0<\alpha\leq \beta$ and $d\in\mathbb{N}/\{0\},M>0$ if:

\begin{enumerate}
  \item $R\geq M+1$,
  \item  $R \geq M+\sqrt{2\beta}\sqrt{d \log\left( d\frac{m^{\nicefrac{2}{d}}\sqrt{\beta}}{\sqrt{\pi\alpha}} \right)}
  = M+\sqrt{2\beta}\sqrt{d \log\left( d\frac{\sqrt{\beta}}{\sqrt{\pi\alpha}} \right) +2\log(m)} $,
  \item $R\geq M+ \sqrt{2\beta}\sqrt{1+\frac{d}{2}}$.
\end{enumerate}

\begin{remark}
\label[remark]{rem: rad_rate}
Note that $R= \mathcal{O}\max(\sqrt{d\log(d)},\sqrt{\log(m)})$ when $R$ is the smallest value satisfying \texttt{Rad}.
\end{remark}
We state a lemma from \cite{panaretos2020invitation} which controls the Wasserstein distance between a measure and its projection on a ball.
\begin{lemma}[Adapted from \cite{panaretos2020invitation}, Equation 2.3]
\label[lemma]{l: wass_proj}
Let $P\in\mathcal{M}_1^+(\mathbb{R}^d)$ and $R>0$. The $1$-Wasserstein distance between $P$ and $\mathcal{P}_R\# P$ is controlled as follows:

\[ W_1(P, \mathcal{P}_R\# P) \leq \int_{||\mathbf{x}||> R} ||\mathbf{x}-\mathcal{P}_R(\mathbf{x})|| dP(\mathbf{x}) \leq \int_{||\mathbf{x}||> R} ||\mathbf{x}|| dP(\mathbf{x}). \]
\end{lemma}
\Cref{l: wass_proj} suggests to consider projected distributions and to control them through the residual moments of the norm of gaussian vectors -- which is done in the following result.

\begin{lemma}
\label[lemma]{l: gaussian_tail}
For $d\geq 3$, $R$ satisfying \texttt{Rad}, any $Q= \mathcal{N}(\mu,\Sigma)\in C_{\alpha,\beta, M}$,
$$Q(||h||> R) \leq \frac{\beta\sqrt{2\beta}}{m}.$$
Also, for any $Q\in C_{\alpha,\beta,M}$:
\[ W_1(Q, \mathcal{P}_R\#Q) \leq \mathbb{E}_{h\sim Q}\left[ ||h|| \mathds{1}(||h||>R) \right] \leq (M+1)\frac{\beta\sqrt{2\beta}}{m}. \]
Finally:
\[ \mathbb{E}_{h\sim Q}\left[ ||h||^2 \mathds{1}(||h||>R) \right] \leq (M+1)^2\frac{\beta\sqrt{2\beta}}{m}. \]
\end{lemma}
The proof of \Cref{l: gaussian_tail} is gathered in \Cref{sec: proof_gaussian_tail}.

\subsection{Differential privacy background}
\label{sec: back_dp}
\begin{definition}[Probability kernels]
A \emph{probability kernel} $\mathcal{P}$ from $\mathcal{Z}^m$ to $\mathcal{M}_1(\mathcal{H})$ is defined as a mapping $\mathcal{P}: {Z}^m \rightarrow \mathcal{M}_1(\mathcal{H})$ .
\end{definition}

\begin{definition}
A probability kernel $\mathcal{P}: \mathcal{Z}^m \rightarrow T$ is $(\varepsilon, \gamma)$-differentially private if, for all pairs $S, S^{\prime} \in Z^m$ that differ at only one coordinate, and all measurable subsets $B \in \Sigma_{\mathcal{H}}$, we have $$\mathbb{P}\{\mathcal{P}(S) \in B\} \leq \mathrm{e}^{\varepsilon} \mathbb{P}\left\{\mathcal{P}\left(S^{\prime}\right) \in B\right\}+\gamma.$$
Further, $\varepsilon$-differentially private means $(\varepsilon, 0)$-differentially private.
\end{definition}

\begin{remark}
Note that classically, differential privacy do not consider stochastic kernels but \emph{randomised algorithms}. Note that this is equivalent to consider probability kernels as precised in \citet[footnote 3, Appendix A]{dziugaite2018data}.
\end{remark}
For our purposes, max-information is the key quantity controlled by differential privacy.
\begin{definition}[\citet{dwork2015gene}, paragraph 3]
Let $\beta \geq 0$, let $X$ and $Y$ be random variables in arbitrary measurable spaces, and let $X^{\prime}$ be independent of $Y$ and equal in distribution to $X$. The \emph{$\beta$-approximate max-information} between $X$ and $Y$, denoted $I_{\infty}^\beta(X ; Y)$, is the least value $k$ such that, for all product-measurable events $E$,
$$
\mathbb{P}\{(X, Y) \in E\} \leq e^k \mathbb{P}\left\{\left(X^{\prime}, Y\right) \in E\right\}+\beta .
$$
The max-information $I_{\infty}(X ; Y)$ is defined to be $I_{\infty}^\beta(X ; Y)$ for $\beta=0$.
For $m \in \mathbb{N}$ and stochastic kernel $\mathcal{P}: \mathcal{Z}^m \rightarrow \mathcal{M}_1(\mathcal{Z})$, the $\beta$-approximate max-information of $\mathcal{P}$, denoted $I_{\infty}^\beta(\mathcal{P}, m)$, is the least value $k$ such that, for all $\mu \in \mathcal{M}_1(\mathcal{Z}), I_{\infty}^\beta(S ; \mathcal{P}(S)) \leq k$
when $S \sim \mathcal{\mu}^m$. The max-information of $\mathcal{P}$ is defined similarly.
\end{definition}
\citet{dziugaite2018data} exploited a boundedness assumption to control the exponential mechanism of \citet{mcsherry2007mechanism}. This ensures that the Gibbs posterior $\mathcal{P}(S)= P_{-\lambda m R_S}$ is $\varepsilon$-diffrentially private for $\varepsilon$ given in \citet[Corollary 5.2]{dziugaite2018data}.
Here, we use a theorem from \citet{minami2016diff} to ensure that for uniformly Lipschitz losses (possibly unbounded), the Gibbs posterior remain $(\varepsilon, \gamma)$-differentially private.

\begin{proposition}[\citet{minami2016diff}, Corollary 8]
\label{prop: minami}
Assume $\mathcal{H}=\mathbb{R}^d$. Assume the loss function to be convex and satisfying \textbf{(A1)}. Finally assume that the (data-free) distribution $P$ is such that $-\log P(.)$ is twice differentiable and $m_P$-strongly convex.
Let $\varepsilon>0, 0<\gamma<1$. Take $\lambda>0$ such that
\[ \lambda \leq \frac{\varepsilon}{2K}\sqrt{\frac{m_P}{1+ 2 \log\left( \nicefrac{1}{\gamma}  \right)}}.  \]
Then the probability kernel $\mathcal{P} : S \rightarrow P_{-\lambda m R_S}$ is $(\varepsilon,\gamma)$-differentially private.
\end{proposition}
Note that, as we mainly focus on Gaussian priors lying on the compact $C_{\alpha,\beta,M}$, the condition on $P$ will always be satisfied with $m_P\geq \alpha$. The last result in this appendix is Theorem 3.1 of \citet{rogers2016max} which upper bounds the $\beta$-approximate max-information of any $(\varepsilon,\gamma)$ differentially private probability kernel.

\begin{proposition}
\label{prop: rogers}
Let $\mathcal{P}: \mathcal{Z}^n \rightarrow \mathcal{M}_1(\mathcal{H})$ be an $(\epsilon, \gamma)$-differentially private probability kernel for $\epsilon \in(0,1 / 2]$ and $\gamma \in(0, \epsilon)$.
For $\beta=e^{-\epsilon^2 m}+O\left(m \sqrt{\frac{\gamma}{\epsilon}}\right)$, we have
$$
I_{\infty}^\beta(\mathcal{P}, m)=O\left(\epsilon^2 m + m \sqrt{\frac{\gamma}{\epsilon}}\right) .
$$
\end{proposition}

\section{Additional proofs}
\label{sec: proofs}
\subsection{Proof of \Cref{th: compact_mcall}}
\label{sec: proof_compact_mcall}

\begin{proof}
We fix $\lambda>0$.

\textit{Step 1: define a good data-dependent function.} We define, for any sample $S$ and predictor $h\in \mathcal{H}$
\[ f_S(h) = \lambda \Delta_S^2(h). \]
This function satisfies the following lemma.
\begin{lemma}
\label[lemma]{l: quasi_lpz_quad}
We fix
$$\varepsilon= \frac{1}{m}, \quad \lambda^{-1}=   K\sqrt{\frac{\log(\frac{N^2}{\delta})}{2m}}\left(\sqrt{\frac{\log(\frac{N^2}{\delta})}{2m}} + 2K\varepsilon \right),$$
with $N= N(\mathcal{H},\varepsilon)$ the $\varepsilon$-covering number of $\mathcal{H}$.
We then have with probability $1-2\delta$ for all $h,h'\in\mathcal{H}$:
\[f_S(h)-f_S(h') \leq \varepsilon_m + ||h-h'||,  \]
with $\varepsilon_m = \frac{4}{\log(\frac{1}{\delta})} \left( 2 + \sqrt{\frac{\log\left(\frac{1}{\delta}\right) + 2d\log(1+2Rm)}{2m}}  \right) = \mathcal{O}\left(1+ \sqrt{\nicefrac{d}{m}}\right)$.
\end{lemma}

\begin{proof}[Proof of \Cref{l: quasi_lpz_quad}]
We rename $N:= N(\mathcal{H},\varepsilon)$.
For any $h,h'\in \mathcal{H}^2$, we have:
\begin{align*}
f_S(h)-f_S(h')  & = \lambda\left(\Delta_S(h)- \Delta_S(h')\right).\left(\Delta_S(h)+ \Delta_S(h')\right).
\end{align*}
The proof of \Cref{l: quasi_lpz_func} gives with probability at least $1-\delta$, for any $h,h'\in \mathcal{H}^2$,
\[\lambda(\Delta_S(h)-\Delta_S(h') \leq 4\lambda K \varepsilon + \sqrt{\frac{\log\left(\frac{N^2}{\delta}\right)}{2m}} \lambda K \left( 2\varepsilon+ ||h-h||\right).  \]
Thus with probability $1-\delta$:
\begin{align*}
f_S(h)-f_S(h') & \leq \left(4\lambda K \varepsilon + \sqrt{\frac{\log\left(\frac{N^2}{\delta}\right)}{2m}} \lambda K \left( 2\varepsilon+ ||h-h||\right) \right).\left(2 \sup_{h\in K} \Delta_S(h)  \right) \\
& = \lambda \left(2K\varepsilon\left( 2 + \sqrt{\frac{\log\left(\frac{N^2}{\delta}\right)}{2m}}  \right) +  K\sqrt{\frac{\log\left(\frac{N^2}{\delta}\right)}{2m}}||h-h'|| \right).\left(2 \sup_{h\in K} \Delta_S(h)  \right).
\end{align*}
Because $\mathcal{H}$ is compact and $\ell$ is $K$-lipschitz, $\Delta_S$ is continuous so there exists $h_S$ such that $\sup_{h\in \mathcal{H}} \Delta_S(h)= \Delta_S(h_S)$.
\medskip

We consider an $\varepsilon$-covering $C:=\{h_1,...,h_N\}$ of $\mathcal{H}$ of size $N$.
Thus, there exists $h_0\in C$ such that $||h_S- h_0||\leq \varepsilon$.
Furthermore, because $\ell \in [0,1]$, by Hoeffding inequality applied for every $h\in C$ and an union bound, we have with probability at least $1-\delta$, for all $h\in C$:
\[ \Delta_S(h) \leq  \sqrt{\frac{\log(\frac{N}{\delta})}{2m}} \leq   \sqrt{\frac{\log(\frac{N^2}{\delta})}{2m}}. \]
Finally using that $\Delta_S$ is $2K$-Lipschitz gives with probability at least $1-\delta$:
\begin{align*}
\sup_{h\in K} \Delta_S(h) &= \Delta_S(h_S) = \Delta_S(h_0) + \left(\Delta_S(h_S)- \Delta_S(h_0)  \right) \\
& \leq  \sqrt{\frac{\log(\frac{N^2}{\delta})}{2m}} + 2K\varepsilon.
\end{align*}
So finally, with probability $1- 2\delta$, we have, for any $h,h'\in \mathcal{H}^2$:
\begin{multline*}
\frac{1}{\lambda} \left(f_S(h)-f_S(h') \right) \\ \leq \left(2K\varepsilon\left( 2 + \sqrt{\frac{\log\left(\frac{N^2}{\delta}\right)}{2m}}  \right) +  K\sqrt{\frac{\log\left(\frac{N^2}{\delta}\right)}{2m}}||h-h'|| \right)\times
2\left( \sqrt{\frac{\log(\frac{N^2}{\delta})}{2m}} + 2K\varepsilon \right).
\end{multline*}
Taking $\lambda^{-1}=  2K\sqrt{\frac{\log(\frac{N^2}{\delta})}{2m}}\left(\sqrt{\frac{\log(\frac{N^2}{\delta})}{2m}} + 2K\varepsilon \right)$ gives:
\begin{align*}
f_S(h)-f_S(h')  & \leq \frac{2\varepsilon\left( 2 + \sqrt{\frac{\log\left(\frac{N^2}{\delta}\right)}{2m}}  \right)}{\sqrt{\frac{\log(\frac{N^2}{\delta})}{2m}}\left(\sqrt{\frac{\log(\frac{N^2}{\delta})}{2m}} + 2K\varepsilon \right)} + ||h-h'||\\
& \leq \frac{4m\varepsilon}{\log\left(\frac{N^2}{\delta}\right)} \left( 2 + \sqrt{\frac{\log\left(\frac{N^2}{\delta}\right)}{2m}}  \right) + ||h-h'|| \\
& \leq \frac{4}{\log(\frac{1}{\delta})} \left( 2 + \sqrt{\frac{\log\left(\frac{1}{\delta}\right) + 2d\log(1+2Rm)}{2m}}  \right) + ||h-h'||.
\end{align*}
The last line holds as $N\geq1$ and that $N\leq N(\bar{\mathcal{B}}(0,R)), \varepsilon) \leq \left(1+\frac{2R}{\varepsilon}\right)^d$ thanks to \Cref{prop: covering} ($\varepsilon = 1/m$).
This proves the lemma.
\end{proof}

\textit{Step 2: A probabilistic change of measure inequality for $f_S$.}
We do not have for the Wasserstein distance such a powerful tool than the change of measure inequality. However, we can generate a probabilistic surrogate on $\mathcal{P}_1(\mathcal{H})$ valid for the function $f_S$ described below.

\begin{lemma}
\label[lemma]{l: change_meas_quad}
For any $\lambda, \varepsilon_m$ defined as in \Cref{l: quasi_lpz_quad}, any $\delta>0$, we have with probability $1-2\delta$ over the sample $S$, for any $P\in\mathcal{P}_1(\mathcal{H})$:

\[ \left(\sup_{Q\in \mathcal{P}_1(\mathcal{H})} \mathbb{E}_{h\sim Q}[ f_S(h)] - \varepsilon_m - W_1(Q,P) \right) \leq \mathbb{E}_{h\sim P}[ f_S(h)].     \]
\end{lemma}

\begin{proof}[Proof of \Cref{l: change_meas_quad}]
For any $\varepsilon>0$, we introduce the cost function $c_{\varepsilon}(x,y)= \varepsilon + ||x-y||$.
From this we notice that we can rewrite the $\varepsilon,1$- Wasserstein distance introduced in \Cref{def: wasserstein} the same way we did in \Cref{l: change_meas}. This leads to
\begin{align*}
W_\varepsilon(Q,P)= \sup_{\substack{(\psi, \phi) \in L^1(Q) \times L^1(P)\\ \psi-\phi \leq c_{\varepsilon}}}\left[\mathbb{E}_{h\sim Q}[ \psi(h)]- \mathbb{E}_{h\sim P}[ \phi(h)]\right].
\end{align*}
A crucial point is that for a well-chosen $\lambda$ with high probability, the pair $(f_S,f_S)$ satisfies the condition stated under the last supremum. It is formalised in the lemma below.
\begin{lemma}
\label{lem:kanto}
Given our choices of $\lambda,\varepsilon_m$, we have with probability at least $1-2\delta$ over the sample $S$ that, for all measures $Q,P\in\mathcal{P}_1(\mathcal{H})^2$:
\begin{itemize}
  \item $f_S\in L_1(Q),L_1(P)$,
  \item for all $h,h' \in \mathcal{H}^2, f_S(h)-f_S(h') \leq c_{\varepsilon_m}(h,h')$.
\end{itemize}
Thus, Kantorovich duality gives us:
\[ \left(\sup_{Q\in \mathcal{P}_1(\mathcal{H})} \mathbb{E}_{h\sim Q}[ f_S(h)] -  W_{\varepsilon_m}(Q,P) \right)\leq \mathbb{E}_{h\sim P}[ f_S(h)],    \]
and using $W_{\varepsilon_m} = \varepsilon_m + W_1$ concludes the proof.
\end{lemma}
\begin{proof}[Proof of \Cref{lem:kanto}]
Because our space of predictors is compact and that for any $z\in\mathcal{Z}$, the loss function $\ell(.,z)$ is $K$-lipschitz on $\mathcal{H}$, then both the generalisation and empirical risk are continuous on $\mathcal{H}$. Thus $|f_S|$ is also continuous and, by compacity, reaches its maximum $M_S$ on $\mathcal{H}$. Thus for any probability $P$ on $K, \mathbb{E}_{h\sim P}[|f_S(h)|] \leq M_S < +\infty$ almost surely. This proves the first statement.
We notice that the second bullet, given our choice of $\lambda$, is the exact conclusion  of \Cref{l: quasi_lpz_quad} with probability at least $1-2\delta$.
So with probability at least $1-2\delta$, Kantorovich duality gives us that for any $P,Q$
\begin{align*}
\mathbb{E}_{h\sim Q}[ f_S(h)] - \mathbb{E}_{h\sim P}[f_S(h)] \leq W_{\varepsilon_m}(Q,P).
\end{align*}
Re-organising the terms and taking the supremum over $Q$ concludes the proof.
\end{proof}
This concludes the proof of \Cref{l: change_meas_quad}.
\end{proof}

\textit{Step 3: The PAC-Bayes proof for the 1-Wasserstein distance.}

We start by exploiting \cref{l: change_meas_quad}: for any prior $P\in\mathcal{P}_1(\mathcal{H})$, for $\lambda, \varepsilon_m$ defined as in \Cref{l: quasi_lpz_quad}, with probability at least $1-2\delta$ we have:

\[ \left(\sup_{Q\in \mathcal{P}_1(\mathcal{H})} \mathbb{E}_{h\sim Q}[ f_S(h)] - \varepsilon_m - W_1(Q,P) \right) \leq \mathbb{E}_{h\sim P}[ f_S(h)]. \]
We then notice that by Jensen's inequality
$$\mathbb{E}_{h\sim P}[ f_S(h)] \leq \frac{\lambda}{2(m-1)}\log\left(\mathbb{E}_{h\sim P}[ \exp(2(m-1)\Delta_S^2(h))]    \right).$$
Then, by Markov's inequality we have with probability $1-\delta$:
\[ \mathbb{E}_{h\sim P}[ f_S(h)] \leq \frac{\lambda}{2(m-1)} \log\left(\frac{\mathbb{E}_S\mathbb{E}_{h\sim P}\left[ \exp\left(2(m-1)\Delta_S^2(h)\right) \right]}{\delta}\right).  \]
By Fubini and Lemma 5 of \citet{mcallester2003simplified}, we have
\[ \mathbb{E}_S\mathbb{E}_{h\sim P}\left[ \exp(f_S(h))\right] \leq m. \]
Taking an union bound and dividing by $\lambda$ gives with probability $1-3\delta$, for any posterior $Q$
\[ \mathbb{E}_{h\sim Q}[\Delta_S^2(h)] \leq \frac{}{}  \frac{W_1(Q,P)+\varepsilon_m}{\lambda} + \frac{\log\left( \frac{m}{\delta} \right)}{2(m-1)}.   \]
We also remark that we can upper bound $\lambda$:
\begin{align*}
\lambda^{-1} & =  2K\sqrt{\frac{\log(\frac{N^2}{\delta})}{2m}}\left(\sqrt{\frac{\log(\frac{N^2}{\delta})}{2m}} + \frac{2K}{m} \right)\\
& \leq 2K(2K+1)\frac{\log(\frac{1}{\delta}) + 2d\log\left(1 +2Rm \right)}{2m}.
\end{align*}
The last line holding because $1/m \leq \sqrt{\frac{\log(\frac{N^2}{\delta})}{2m}}$. Also $N= N(\mathcal{H},1/m) \leq (1+2Rm)^d$ thanks to \cref{prop: covering}.
Then, bounding $1/2m, 1/(2m-1)$ by $1/m$ gives, with probability at least $1-3\delta$, for any posterior $Q$
\begin{align*}
\mathbb{E}_{h\sim Q}[\Delta_S^2(h)] & \leq  2K(2K+1) \frac{\log(\frac{1}{\delta}) + 2d\log\left(1 +2Rm \right)}{m} \left(W_1(Q,P)+\varepsilon_m \right) + \frac{\log\left( \frac{m}{\delta} \right)}{m}.
\end{align*}
We finally exploit Jensen's inequality once more to remark that for any $Q$, $\mathbb{E}_{h\sim Q}[\Delta_S^2(h)] \geq \left(\mathbb{E}_{h\sim Q}[\Delta_S(h)]  \right)^2$.
Then, with probability at least $1-3\delta$, for any posterior $Q$
\[ |\Delta_S(Q)| \leq \sqrt{2K(2K+1)\frac{2d\log\left(\frac{1 +2Rm }{\delta}\right)}{m} \left(W_1(Q,P)+\varepsilon_m \right) + \frac{\log\left( \frac{m}{\delta} \right)}{m}   } \]
Taking $\delta'= \delta/3$ concludes the proof.
\end{proof}

\subsection{Proof of \Cref{l: gaussian_tail}}
\label{sec: proof_gaussian_tail}

\begin{proof}[Proof of \Cref{l: gaussian_tail}]
We denote by $\mathbf{x}$ a vector of $\mathbb{R}^d$, by $d\mathbf{x}= d_{x_1}...dx_{d}$ the Lebesgue measure on $\mathbb{R}^d$ and $f_{\mu,\Sigma}(\mathbf{x})= \exp\left(\frac{1}{2}(\mathbf{x}^T-m) \Sigma^{-1} (\mathbf{x}-m)  \right)$.

\textit{First bound.}
First we use that $||\mu||\leq M$ to say that $\bar{\mathcal{B}}(0_{\mathbb{R}^d}, R-M) \subseteq \bar{\mathcal{B}}(-m, R) $ and so:
\begin{align*}
\sqrt{(2 \pi)^d|\Sigma|}.Q(||x||> R) & = \int_{||\mathbf{x}||> R} f_{\mu,\Sigma}(\mathbf{x})d\mathbf{x} \leq  \int_{||\mathbf{x}||> R- M} f_{0,\Sigma}(\mathbf{x})d\mathbf{x}
\end{align*}
where $ |\Sigma|$ the determinant of $\Sigma$.
We now use that because $Q\in C_{\alpha,\beta, M}, \alpha Id \preceq \Sigma \preceq \beta Id$. We then have: $|\Sigma|\geq \alpha^d$ and for any $\mathbf{x}$, $\mathbf{x}^T \Sigma^{-1} \mathbf{x} \geq ||\mathbf{x}||^2/\beta.$
Thus we have:
\begin{align*}
Q(||h||>R) & = \frac{1}{\sqrt{2 \pi\alpha}^{d}}\int_{||\mathbf{x}||>(R-M)} \exp\left(\frac{1}{2\beta}||\mathbf{x}||^2  \right) d \mathbf{x}
\end{align*}
We use the hyperspherical coordinate (see \emph{e.g.} \citealp{spherical_coord}) to obtain:
\begin{align*}
\int_{||\mathbf{x}||>(R-M)} \exp\left(\frac{1}{2\beta}||\mathbf{x}||^2  \right) d \mathbf{x} &= \int_{R-M}^{+\infty} r^{d-1} \exp\left(- \frac{r^2}{2\beta}\right)dr\\
& \leq \int_{R-M}^{+\infty} r^{d+1} \exp\left(- \frac{r^2}{2\beta}\right)dr \\
&= \beta\sqrt{2\beta}^{d+1} \int_{\frac{(R-M)^2}{2\beta}}^{+\infty} r^{\frac{d}{2}} \exp^{-r}dr.\\
\end{align*}
The second line holding because we assumed $R-M\geq 1$ thanks to \texttt{Rad}. We define the \emph{residual of Euler's Gamma function} as:   $\Gamma\left(1+\frac{d}{2}, \frac{(R-M)^2}{2\beta}\right):= \int_{\frac{(R-M)^2}{2\beta}}^{+\infty} r^{\frac{d}{2}} \exp^{-r}dr$.
Then we use \citet[Lemma 4.4.3, p.84]{gabcke1979neue} which ensure us that (because point 3 of \texttt{Rad} gives $\frac{(R-M)^2}{2\beta}\geq 1+\frac{d}{2}$):
\begin{align*}
\Gamma\left(1+\frac{d}{2}, \frac{(R-M)^2}{2\beta}\right) & \leq \frac{d+2}{2}\exp\left({-\frac{(R-M)^2}{2\beta}} \right) \left(\frac{(R-M)^2}{2\beta}\right)^{\nicefrac{d}{2}}.
\end{align*}
We now control this quantity through the following lemma.
\begin{lemma}
\label{l: calculus}
Let $d\geq 3$, $f(r)= \frac{d}{2}\log(r) -r$ Then for any $r=\frac{(R-M)^2}{2\beta}$ with $R$ satisfying \texttt{Rad}, we have :
\[f(r) \leq -\frac{d}{2}\log\left(\sqrt{\frac{\beta}{\pi\alpha}}\right) - \log(m)- \log\left(\frac{d+2}{2}\right).\]
\end{lemma}
\begin{proof}[Proof of \Cref{l: calculus}]
First of all, $f$ is decreasing on $[\nicefrac{d}{2},+ \infty).$ Notice that if $r_0= d \log\left( d\frac{m^{\nicefrac{2}{d}}\sqrt{\beta}}{\sqrt{\pi\alpha}} \right)$, then $r_0\geq \nicefrac{d}{2}$ because $d\geq 3$.
Thus, $r=\frac{(R-M)^2}{2\beta}$, with $R$ satisfying \texttt{Rad}. We then know that $r\geq r_0$ so $f(r)\leq f(r_0)$.
The only thing left to prove is that
\[f(r_0)\leq -\frac{d}{2} \log\left(\sqrt{\frac{\beta}{\pi\alpha}}\right) - \log(m) -  \log\left(\frac{d+2}{2}\right). \]
To do so, notice that:
\begin{align*}
\log(r_0)& = \log(d)+ \log\left(\log\left(dm^{\nicefrac{2}{d}}\sqrt{\frac{\beta}{\pi \alpha}}  \right)  \right).
\intertext{So, multiplying by $d/2$ gives:}
\frac{d}{2}\log(r_0)&= -\frac{d}{2}\log\left(m^{\nicefrac{2}{d}} \sqrt{\frac{\beta}{\pi \alpha}} \right)
+ \frac{r_0}{2} + \frac{d}{2}\log\log\left( dm^{\nicefrac{2}{d}}\sqrt{\frac{\beta}{\pi \alpha}} \right).
\intertext{Finally:}
f(r_0) &= -\frac{d}{2}\log\left( m^{\nicefrac{2}{d}}\sqrt{\frac{\beta}{\pi \alpha}} \right) - \frac{r_0}{2}
+ \frac{d}{2}\log\log\left( dm^{\nicefrac{2}{d}}\frac{\beta}{\pi \alpha} \right)
\end{align*}
We conclude the proof by proving $$- \frac{r_0}{2} + \frac{d}{2}\log\log\left( dm^{\nicefrac{2}{d}}\sqrt{\frac{\beta}{\pi \alpha}} \right) \leq -\log\left( \frac{d+2}{2} \right).$$
Note that this is equivalent to
\[ dm^{\nicefrac{2}{d}}\sqrt{\frac{\beta}{\pi \alpha}} -  \left(1+ \frac{d}{2}\right)^{\nicefrac{2}{d}}\log\left( dm^{\nicefrac{2}{d}}\sqrt{\frac{\beta}{\pi \alpha}} \right) \geq 0. \]
This is true because for $d\geq 3$, $\left(1+ \frac{d}{2}\right)^{\nicefrac{2}{d}} \leq 2$ and the function $\mathbb{R}^+$, $x\rightarrow x - 2\log(x)$ is positive. This concludes the proof.
\end{proof}
We then have
\[ \exp\left({-\frac{(R-M)^2}{2\beta}} \right) \left(\frac{(R-M)^2}{2\beta}\right)^{\nicefrac{d}{2}} = \exp\left(f \left( \frac{(R-M)^2}{2\beta} \right)\right) \leq
\sqrt{\frac{\pi\alpha}{\beta}}^{d} \times \frac{2}{d+2} \times \frac{1}{m} . \]
Hence the final bound:
\[ Q(||h||>R)  \leq  \frac{\beta\sqrt{2\beta}}{m} .  \]

\textit{Second bound.}
We use \cref{l: wass_proj} to have
\begin{align*}
W_1(Q, \mathcal{P}_R\# Q) & \leq \int_{||\mathbf{x}||> R} ||\mathbf{x}- \mathcal{P}_R(\mathbf{x})|| dP(\mathbf{x}).
\intertext{By definition of the projection on a closed convex, $||\mathbf{x}-\mathcal{P}_R(\mathbf{x})|| \leq ||\mathbf{x}||$. Thus:}
&  \leq  \frac{1}{\sqrt{(2 \pi)^d|\Sigma|}}\int_{||\mathbf{x}||> R} ||\mathbf{x}||f_{\mu,\Sigma}(\mathbf{x})d\mathbf{x} \\
& \leq \frac{1}{\sqrt{(2 \pi)^d|\Sigma|}}\int_{||\mathbf{x}||> R} ||\mathbf{x}- \mu||f_{\mu,\Sigma}(\mathbf{x})d\mathbf{x} + MQ(||h||>R) \\
& \leq \frac{1}{\sqrt{(2 \pi)^d|\Sigma|}} \int_{||\mathbf{x}||> R} ||\mathbf{x}-\mu||f_{\mu,\Sigma}(\mathbf{x})d\mathbf{x} + \frac{M\beta\sqrt{2\beta}}{m}.
\intertext{The last line holding thanks to the first part of the proof, then using again that $||\mu||\leq M$ gives:}
W_1(Q, \mathcal{P}_R\# Q) & \leq\frac{1}{\sqrt{(2 \pi)^d|\Sigma|}} \int_{||\mathbf{x}||> R-M} ||\mathbf{x}||f_{0,\Sigma}(\mathbf{x})d\mathbf{x} + \frac{M\beta\sqrt{2\beta}}{m}.
\intertext{Then using the same arguments than in the first part of the proof gives:}
W_1(Q, \mathcal{P}_R\# Q) & \leq \frac{1}{\sqrt{2 \pi\alpha}^{d}} \int_{||\mathbf{x}||> R-M} ||\mathbf{x}|| \exp\left( -\frac{||\mathbf{x}||^2}{2\beta} \right)d\mathbf{x} + \frac{M\beta\sqrt{2\beta}}{m}.
\end{align*}
We use the hyperspherical coordinate to obtain:
\begin{align*}
\int_{||\mathbf{x}||> R-M} ||\mathbf{x}|| \exp\left( -\frac{||\mathbf{x}||^2}{2\beta} \right)d\mathbf{x} &= \int_{R-M}^{+\infty} r^d \exp\left(- \frac{r^2}{2\beta}\right)dr\\
& \leq \int_{R-M}^{+\infty} r^{d+1} \exp\left(- \frac{r^2}{2\beta}\right)dr \\
&= \beta\sqrt{2\beta}^{d+1} \int_{\frac{(R-M)^2}{2\beta}}^{+\infty} r^{\nicefrac{d}{2}} \exp^{-r}dr\\
& = \beta\sqrt{2\beta}^{d+1}\Gamma\left(\frac{d+1}{2}, \frac{(R-M)^2}{2\beta}\right).
\end{align*}
The second line holding because $R-M\geq 1$.
Then applying again \Cref{l: calculus} gives:
\begin{align*}
\mathbb{E}_{h\sim Q}\left[ ||h|| \mathds{1}(||h||>R) \right]& \leq  \beta\sqrt{2\beta} \sqrt{\frac{\beta}{\pi\alpha}}^d\times \frac{d+2}{2} \sqrt{\frac{\pi\alpha}{\beta}}^{d} \times \frac{2}{d+2} \times \frac{1}{m}
+ \frac{M\beta\sqrt{2\beta}}{m} \\
& = (M+1)\frac{\beta\sqrt{2\beta}}{m}.
\end{align*}
Hence the final bound:
\[ W_1(Q, \mathcal{P}_R\# Q)  \leq  \mathbb{E}_{h\sim Q}\left[ ||h|| \mathds{1}(||h||>R) \right] \leq (M+1)\frac{\beta\sqrt{2\beta}}{m}.  \]

\textit{Third bound.}
We start again as
\begin{align*}
\mathbb{E}_{h\sim Q}[||h||^2 \mathds{1}(||h||>R)] & =  \frac{1}{\sqrt{(2 \pi)^d|\Sigma|}}\int_{||\mathbf{x}||> R} ||\mathbf{x}||^2f_{\mu,\Sigma}(\mathbf{x})d\mathbf{x} \\
& =\frac{1}{\sqrt{(2 \pi)^d|\Sigma|}}\int_{||\mathbf{x}||> R} ||\mathbf{x}- \mu||^2 +2\langle \mu, \mathbf{x}- \mu\rangle + ||\mu||^2 f_{\mu,\Sigma}(\mathbf{x})d\mathbf{x}.
\intertext{Then, using that $\mu$ is the mean of $Q$ and that $||\mu||\leq M$ gives: }
\mathbb{E}_{h\sim Q}[||h||^2 \mathds{1}(||h||>R)] & \leq \frac{1}{\sqrt{(2 \pi)^d|\Sigma|}} \int_{||\mathbf{x}||> R} ||\mathbf{x}-\mu||^2f_{\mu,\Sigma}(\mathbf{x})d\mathbf{x}\\
& + 2M \frac{1}{\sqrt{(2 \pi)^d|\Sigma|}} \int_{||\mathbf{x}||> R} ||\mathbf{x}-\mu||f_{\mu,\Sigma}(\mathbf{x})d\mathbf{x} + M^2Q(||h||>R).
\intertext{Then, the first and second bounds of \cref{l: gaussian_tail} give}
\mathbb{E}_{h\sim Q}[||h||^2 \mathds{1}(||h||>R)] & \leq \frac{1}{\sqrt{(2 \pi)^d|\Sigma|}} \int_{||\mathbf{x}||> R} ||\mathbf{x}-\mu||^2f_{\mu,\Sigma}(\mathbf{x})d\mathbf{x} + (M^2 + 2M) \frac{\beta\sqrt{2\beta}}{m}.
\intertext{Finally:}
\mathbb{E}_{h\sim Q}[||h||^2 \mathds{1}(||h||>R)] & \leq \frac{1}{\sqrt{2 \pi\alpha}^{d}} \int_{||\mathbf{x}||> R-M} ||\mathbf{x}||^2 \exp\left( -\frac{||\mathbf{x}||^2}{2\beta} \right)d\mathbf{x} + (M^2 + 2M +2)\frac{\beta\sqrt{2\beta}}{m}.
\end{align*}
We use the hyperspherical coordinate to obtain:
\begin{align*}
\int_{||\mathbf{x}||> R-M} ||\mathbf{x}||^2 \exp\left( -\frac{||\mathbf{x}||^2}{2\beta} \right)d\mathbf{x} &= \int_{R-M}^{+\infty} r^{d+1} \exp\left(- \frac{r^2}{2\beta}\right)dr\\
&= \beta\sqrt{2\beta}^{d+1} \int_{\frac{(R-M)^2}{2\beta}}^{+\infty} r^{\nicefrac{d}{2}} \exp^{-r}dr\\
& = \beta\sqrt{2\beta}^{d+1}\Gamma\left(\frac{d+1}{2}, \frac{(R-M)^2}{2\beta}\right).
\end{align*}
Then applying again \Cref{l: calculus} gives:
\begin{align*}
\mathbb{E}_{h\sim Q}\left[ ||h||^2 \mathds{1}(||h||>R) \right]& \leq  \frac{\beta\sqrt{2\beta}}{m}
+ (M^2 + 2M)\frac{\beta\sqrt{2\beta}}{m} \\
& = (M+1)^2\frac{\beta\sqrt{2\beta}}{m}.
\end{align*}
This concludes the proof.

\end{proof}

\subsection{Proof of \Cref{th: main_gaussian_smooth}}
\label{sec: proof_smooth}

\begin{proof}[Proof of \Cref{th: main_gaussian_smooth}]
We take a specific radius $R$ which is the smallest value satisfying \texttt{Rad}.
We first notice that because for all $z$, $\ell(.,z)$ is $L$-smooth, then on $\mathcal{B}(0,R)$, the gradients of $\ell(.,z)$ are bounded by $D_R=D + LR$. Thus $\ell$ is uniformly $D_R$-Lipschitz on the closed ball of radius $R$.
This allow us a straightforward application of \Cref{th: compact_mcall} on the compact $\mathcal{B}(0,R)$, with the prior $\mathcal{P}_R\# P$, and with high probability, for any posterior $\mathcal{P}_R\# Q$ with $Q\in C_{\alpha,\beta,M}$:
\begin{align*}
|\Delta_S(\mathcal{P}_R\# Q)|  \leq  \sqrt{2D_R(2D_R+1)\frac{2d\log\left(3\frac{1 +2Rm }{\delta}\right)}{m} \left(W_1(\mathcal{P}_R\# Q,\mathcal{P}_R\# P) + \varepsilon_m \right) + \frac{\log\left( \frac{3m}{\delta} \right)}{m} }.
\end{align*}
From this we control the left hand-side term as follows:
\begin{align*}
|\Delta_S(Q)| &\leq |\Delta_S(\mathcal{P}_R\# Q)| + |\Delta_S(Q) - \Delta_S(\mathcal{P}_R\# Q) |
\intertext{And we also have as in the proof of \Cref{th: main_gaussian_lpz}:}
|\Delta_S(Q) - \Delta_S(\mathcal{P}_R\# Q) | & \leq 2Q(||h||>R) \leq 2 \frac{\beta\sqrt{2\beta}}{m}.
\end{align*}
Also we have by the triangle inequality:
\[W_1(\mathcal{P}_R\# Q,\mathcal{P}_R\# P) \leq W_1(Q, \mathcal{P}_R\# Q) + W_1(Q,P)+ W_1(P, \mathcal{P}_R\# P). \]
Because both $Q,P\in C_{\alpha,\beta,M}$, using again \Cref{l: gaussian_tail} gives:
\[W_1(\mathcal{P}_R\# Q,\mathcal{P}_R\# P) \leq  W_1(Q,P)+2(M+1)\frac{\beta\sqrt{2\beta}}{m}. \]
We then have:
\begin{align*}
|\Delta_S(Q)| \leq 2 \frac{\beta\sqrt{2\beta}}{m} + \sqrt{2D_R(2D_R+1)\frac{2d\log\left(3\frac{1 +2Rm }{\delta}\right)}{m} \left(W_1(Q,P)+ \alpha_m \right) + \frac{\log\left( \frac{3m}{\delta} \right)}{m} }.
\end{align*}
with $\alpha_m= 2(M+1)\frac{\beta\sqrt{\beta}}{m} + \varepsilon_m= \mathcal{O}\left(1 + \sqrt{\nicefrac{d\log(Rm)}{m}}\right)$. This concludes the proof.
\end{proof}